\documentclass{article}

\usepackage{arxiv}
\usepackage[utf8]{inputenc} 
\usepackage[T1]{fontenc}    
\usepackage{hyperref}       
\usepackage{url}            
\usepackage{booktabs}       
\usepackage{amsfonts}       
\usepackage{nicefrac}       
\usepackage{microtype}      
\usepackage{subfig}
\usepackage{lipsum}
\usepackage{caption}
\usepackage{graphics}
\usepackage{graphicx}
\usepackage{algorithm2e}
\usepackage{cite}
\usepackage{comment}
\usepackage{float}
\usepackage{amsmath}
\usepackage{multirow}
\usepackage[dvipsnames]{xcolor}
\usepackage[title]{appendix}
\usepackage{ulem}
\usepackage{cancel}
\usepackage{tabularx}
\usepackage{colortbl}

\usepackage{xparse}

\usepackage{lineno}
\usepackage[onehalfspacing]{setspace}

\usepackage{amsmath}

\DeclareMathOperator*{\argmax}{\arg\!\max}
\DeclareMathOperator*{\argmin}{\arg\!\min}
\DeclareMathOperator*{\MMD}{MMD}

\DeclareMathOperator*{\Dis}{DPR}
\DeclareMathOperator*{\NN}{NN}
\DeclareMathOperator*{\proj}{proj}

\title{Training and Evaluation of Deep Policies using Reinforcement Learning and Generative Models}

\author{
  \thanks{denotes an equal contribution} Ali Ghadirzadeh \thanks{Stanford University} \\
  \And
  $^*$Petra Poklukar \thanks{KTH Royal Institute of Technology, Division of Robotics, Perception and Learning} \\
  \And
  Karol Arndt\,\thanks{Aalto University, Intelligent Robotics Research Group} \\
  \AND
  Chelsea Finn$^{\dagger}$ \\
  \And 
  Ville Kyrki\,$^{\S}$ \\
  \And
  Danica Kragic\,$^{\ddagger}$ \\
  \And
  M\aa rten Bj\"orkman\,$^{\ddagger}$ \\
}

\begin{document}
\maketitle

\begin{abstract}
We present a data-efficient framework for solving sequential decision-making problems which exploits the combination of reinforcement learning (RL) and latent variable generative models. The framework, called GenRL, trains deep  policies by introducing an action latent variable such that the feed-forward policy search can be divided into two parts: (i) training a sub-policy that outputs a distribution over the action latent variable given a state of the system, and (ii) unsupervised training of a generative model that outputs a sequence of motor actions conditioned on the latent action variable.
GenRL enables safe exploration and alleviates the data-inefficiency problem as it exploits prior knowledge about valid sequences of motor actions. Moreover, we provide a set of measures for evaluation of generative models such that we are able to predict the performance of the RL policy training prior to the actual training on a physical robot. 
We experimentally determine the characteristics of generative models that have most influence on the performance of the final policy training on two robotics tasks: shooting a hockey puck and throwing a basketball. Furthermore, we empirically demonstrate that GenRL is the only method which can safely and efficiently solve the robotics tasks compared to two state-of-the-art RL methods. 
\end{abstract}
\section{Introduction}
\label{sec:introduction}

Reinforcement learning (RL) can leverage modeling capability of generative models to solve complex sequential decision making problems more efficiently \cite{singh2020parrot, ghadirzadeh2017deep, arndt2019meta}. RL has been applied to end-to-end training of deep visuomotor robotic policies \cite{levine2016end,levine2018learning} but it is typically too data-inefficient especially when applied to tasks that provide only a terminal reward at the end of an episode. One way to alleviate the data-inefficiency problem in RL is by leveraging prior knowledge to reduce the complexity of the optimization problem. One prior that significantly reduces the data requirement is an approximation of the distribution from which valid action sequences can be sampled. Such distributions can be efficiently approximated by training generative models given a sufficient amount of valid action sequences. 

The question is then how to combine powerful RL optimization algorithms with the modeling capability of generative models to improve the efficiency of the policy training? Moreover, which characteristics of the generative models are important for efficient policy training? A suitable generative model must capture the entire distribution of the training data to generate as many distinct motion trajectories as possible, while avoiding the generation of invalid trajectories outside the training dataset. 
The diversity of the generated data enables the policy to complete a given task for the entire set of goal states when training a goal-conditioned policy. On the other hand, adhering to the distribution of the training data ensures safety in generated trajectories which are running on a real robotic platform. 

In this paper, we (i) propose a learning framework that exploits RL and generative models to solve sequential decision making problems and considerably improves the data-efficiency of deep policy training, and (ii) provide a set of measures to evaluate the quality of the latent space of different generative models regulated by the RL policy search algorithms, and use them as a guideline for the training of generative models such that the data-efficiency of the policy training can be further improved prior to actual training on a physical robot. 

Regarding (i), we propose the \textit{GenRL} learning framework that divides the sequential decision-making problem into the following sub-problems which can be solved more efficiently:
(1) an unsupervised generative model training problem that approximates the distribution of motor actions, and (2) a trust-region policy optimization problem that solves a contextual multi-armed bandit without temporal credit assignment issue which exists in typical sequential decision-making problems. 

Regarding (ii), we evaluate generative models based on (a) the quality and coverage of the samples they generate using the precision and recall metric~\cite{kynkaanniemi2019improved}, and 
(b) the quality of their latent representations by measuring their disentanglement using a novel measure called \textit{disentanglement with precision and recall (DwPR)} which is derived from the above precision and recall metric. The measures can be applied to any task prior to the policy training as they do not leverage the end states obtained after execution of the generated trajectories on a robotic platform.
Our hypothesis is that a generative model which is able to generate realistic samples that closely follow the training data (i.e. has high precision and high recall) and is well disentangled (i.e. has a latent space where individual dimensions control different aspects of the task) leads to a more sample efficient neural network policy training. 

We experimentally evaluate the GenRL framework on two robotics tasks: shooting a hockey puck and throwing a basketball. 
To quantify the importance of (a) and (b) for a superior RL policy training performance and validate our hypothesis, we perform a detailed evaluation of several generative models, namely $\beta$-VAEs \cite{higgins2017beta} and InfoGANs \cite{chen2016infogan}, by calculating Pearson's R between their characteristics and the performance of the policy. Furthermore, we demonstrate the safety and data-efficiency of GenRL on the two tasks by comparing it to two state-of-the-art policy search algorithms, PPO \cite{schulman2017proximal} and SAC \cite{haarnoja2018soft}.
In summary, the advantages of the proposed GenRL framework are: 
\begin{itemize}
    \item It enables learning complex robotics skills such as throwing a ball by incorporating prior knowledge in terms of a distribution over valid sequences of actions, therefore, reducing the search space  (Section~\ref{sec:exp:data-efficiency}).  
    
    \item It helps to learn from sparse terminal rewards that are only provided  at the end of successful episodes. The proposed formulation converts the sequential decision-making problem into a contextual multi-armed bandit (Section~\ref{sec:preliminaries}). Therefore, it alleviates the temporal credit assignment problem that is inherent in  sequential decision-making tasks and enables efficient policy training with only terminal rewards. 
    
    \item It enables safe exploration in RL by sampling actions only from the approximated distribution (Section~\ref{sec:exp:safety}). 
    This is in stark contrast to the typical RL algorithms in which random actions are taken during the exploration phase. 
    
    \item It provides a set of measures for evaluation of the generative model based on which it is possible to predict the performance of the RL policy training prior to the actual training (Section~\ref{sec:exp:eval_gen_models}).
\end{itemize}

This paper provides a comprehensive overview of our earlier work for RL policy training based on generative models~\cite{ghadirzadeh2017deep, arndt2019meta, chen2019adversarial,hamalainen2019affordance, butepage2019imitating} and is organized as follows: in Section \ref{sec:related_work}, we provide an overview of the related work. We formally introduce the problem of policy training with generative models in Section~\ref{sec:preliminaries}, and describe how the GenRL framework is trained in Section~\ref{sec:em_policy_training}. In Section~\ref{sec:generative_model_training} we first briefly overview $\beta$-VAEs and InfoGANs, and then define the evaluation measures used to predict the final policy training performance.  
We present the experimental results in Section \ref{sec:experiments} and discuss the conclusion and future work in Section~\ref{sec:conclusion}.
Moreover, for the sake of completeness, we describe the end-to-end training of the perception and control modules in Appendix \ref{app:perception} by giving a summary of the prior work \cite{levine2016end, chen2019adversarial}. Note that this work provides a complete overview of the proposed GenRL framework and focuses more on the evaluation of the generative model. We refer the reader to \cite{ghadirzadeh2017deep} for further investigation of the data-efficiency of the proposed approach in training complex visuomotor skills.
Our code will be made publicly available upon acceptance.

\section{Related work}
\label{sec:related_work}

Our work addresses two types of problems: (a) policy training based on unsupervised generative model training and trust-region policy optimization, and (b) evaluation of generative models to forecast the efficiency of the final policy training task. 
We introduce the related work for each of the problems in the following sections. 

\textbf{Data-efficient end-to-end policy training}
In recent years, end-to-end training of policies using deep RL has gained in popularity in robotics research \cite{ghadirzadeh2017deep, levine2016end, finn2016deep, kalashnikov2018qt, quillen2018deep, singh2017gplac, devin2018deep, pinto2017asymmetric}. 
However, deep RL algorithms are typically data-hungry and learning a general policy, i.e., a policy that performs well also for previously unseen inputs, requires a farm of robots continuously collecting data for several days \cite{levine2018learning, finn2017deep, gu2017deep, dasari2019robonet}.
The limitation of large-scale data collection has hindered the applicability of RL solutions to many practical robotics tasks. Recent studies tried to improve the data-efficiency by training the policy in simulation and transferring the acquired  skills to the real setup \cite{quillen2018deep, pinto2017asymmetric, abdolmaleki2020distributional, peng2018sim}, a paradigm known as sim-to-real transfer learning. Sim-to-real approaches are utilized for two tasks in deep policy training: (i) training the perception model via randomization of the texture and shape of visual objects in simulation and using the trained model directly in the real world setup (zero-shot transfer) \cite{hamalainen2019affordance, tobin2017domain}, and (ii) training the policy in simulation by randomizing the dynamics of the task and transferring the policy to the real setup by fine-tuning it with the real data (few-shot transfer learning) \cite{arndt2019meta, peng2018sim}. However, challenges in the design of the simulation environment can cause large differences between the real and the simulated environments which hinder an efficient knowledge transfer between these two domains. 
In such cases, transfer learning from other domains, e.g., human demonstrations \cite{butepage2019imitating, yu2018one} or simpler task setups \cite{chen2019adversarial, chen2018deep}, can help the agent to learn a policy more efficiently. 
In this work, we exploit human demonstrations to shape the robot motion trajectories by training generative models that reproduce the demonstrated trajectories. 
Following our earlier work \cite{chen2019adversarial}, we exploit adversarial domain adaptation techniques \cite{tzeng2017adversarial, tzeng2020adapting} to improve the generality of the acquired policy when it is trained in a simple task environment with a small amount of training data. In the rest of this section, we review related studies that improve the data-efficiency and generality of RL algorithms by utilizing trust-region terms, converting the RL problem into a supervised learning problem, and trajectory-centered approaches that shape motion trajectories prior to the policy training.

Improving the policy while avoiding abrupt changes in the policy distribution after each update is known as the trust-region approach in policy optimization. 
Trust-region policy optimization (TRPO) \cite{schulman2015trust} and proximal policy optimization (PPO) \cite{schulman2017proximal} are two variants of trust-region policy gradient methods that scale well to non-linear policies such as neural networks. The key component of TRPO and PPO is a surrogate objective function with a trust-region term based on which the policy can be updated and monotonically improved. In TRPO, the changes in the distributions of the policies before and after each update are penalized by a KL divergence term. 
Therefore, the policy is forced to stay in a trust-region given by the action distribution of the current policy. Our expectation-maximization (EM) formulation yields a similar trust-region term with the difference being that it penalizes the changes in the distribution of the deep policy and a so-called variational policy that will be introduced as a part of our proposed optimization algorithm. Since our formulation allows the use of any policy gradient solution, we use the same RL objective function as in TRPO. 

The EM algorithm has been used for policy training in a number of prior work \cite{neumann2011variational,deisenroth2013survey,levine2013variational}. The key idea is to introduce variational policies to decompose the policy training into two downstream tasks that are trained iteratively until no further policy improvement can be observed \cite{ghadirzadeh2018sensorimotor}. 
In \cite{levine2016end} the authors introduced the guided policy search (GPS) algorithm which divides the visuomotor policy training task into a trajectory optimization and a supervised learning problem. 
GPS alternates between two steps: (i) optimizing a set of trajectories by exploiting a trust-region term to stay close to the action distribution given by the deep policy, and (ii) updating the deep policy to reproduce the motion trajectories. 
Our EM solution differs from the GPS framework and earlier approaches in that we optimize the trajectories by regulating a generative model that is trained prior to the policy training. Training generative models enables the learning framework to exploit human expert knowledge as well as to optimize the policy given only terminal rewards as explained earlier.

Trajectory-centric approaches, such as dynamic movement primitives (DMPs), have been popular because of the ease of integrating expert knowledge in the policy training process via physical demonstration \cite{bahl2020neural, peters2006policy, peters2008reinforcement, ijspeert2003learning, ijspeert2013dynamical, hazara2019transferring}.
However, such models are less expressive compared to deep neural networks. Moreover, these approaches cannot be used to train reactive policies where the action is adjusted in every time-step based on the observed sensory input \cite{haarnoja2018composable}. 
On the other hand, deep generative models can model complex dependencies within the data by learning the underlying data distribution from which realistic samples can be obtained. 
Furthermore, they can be easily accommodated in larger neural networks without affecting the data integrity. Our GenRL framework based on generative models enables training both feedback (reactive) and feedforward policies by adjusting the policy network architecture. 

The use of generative models in robot learning has become popular in recent years \cite{zhou2020plas, chen2022latent, ghadirzadeh2017deep, butepage2019imitating, hamalainen2019affordance, chen2019adversarial, arndt2019meta, lippi2020latent, gothoskar2020learning, igl2018deep, buesing2018learning, mishra2017prediction, ke2018modeling, hafner2018learning, rhinehart2018deep, krupnik2019multi} because of their  low-dimensional and regularized latent spaces. However, latent variable generative models are mainly studied to train a long-term state prediction model that is used in the context of trajectory optimization and model-based reinforcement learning  \cite{buesing2018learning, mishra2017prediction, ke2018modeling, hafner2018learning, rhinehart2018deep, krupnik2019multi}.
Regulating generative models based on reinforcement learning to produce sequences of actions according to the  visual state has first appeared in our prior work \cite{ghadirzadeh2017deep}. Since then we applied the framework in different robotic tasks, e.g., throwing balls \cite{ghadirzadeh2017deep}, shooting hockey-pucks \cite{arndt2019meta}, pouring into mugs \cite{chen2019adversarial, hamalainen2019affordance}, and in a variety of problem domains, e.g., sim-to-real transfer learning \cite{hamalainen2019affordance, arndt2019meta} and domain adaptation to acquire general policies \cite{chen2019adversarial}. 
Recently, generative models have also been used to train feedback RL policies based on  normalizing flows \cite{dinh2016density} to learn behavioral prior that transforms the original MDP into a simpler one for the RL policy training 
\cite{singh2020parrot}. 
In this work, we focus on RL tasks that can be best implemented using feed-forward policies, and study the properties of generative models that improve the efficiency of the RL policy training. 

\textbf{Evaluation of generative models}
Although generative models have proved successful in many domains \cite{lippi2020latent, brock2018large, wang2018high, vae_anom, vae_text8672806} assessing their quality remains a challenging problem \cite{challenging_common, poklukar2022delaunay, pmlr-v139-poklukar21a}. It involves analysing the quality of both latent representations and generated samples. Regarding the latter, generated samples and their diversity should resemble those obtained from the training data distribution. Early developed metrics such as IS \cite{IS_NIPS2016_6125}, FID \cite{FID_NIPS2017_7240} and KID \cite{binkowski2018demystifying} provided a promising start but were shown to be unable to separate between failure cases, such as mode collapse or unrealistic generated samples \cite{sajjadi2018assessing, kynkaanniemi2019improved}. Instead of using a one-dimensional score, \cite{sajjadi2018assessing} proposed to evaluate the learned distribution by comparing the samples from it with the ground truth training samples using the notion of precision and recall. Intuitively, precision measures the similarity between the generated and real samples, while recall determines the fraction of the true distribution that is covered by the distribution learned by the model. The measure was further improved both theoretically and practically by \cite{revisiting_pr}, while \cite{kynkaanniemi2019improved} provides an explicit non-parametric variant of the original probabilistic approach. We use the precision and recall measure provided by \cite{kynkaanniemi2019improved}. 

Regarding the assessment of the quality of the latent representation,
a widely adopted approach is the measure of disentanglement \cite{higgins2018towards, repr_learning_survey, tschannen2018recent}. A representation is said to be disentangled if each latent component encodes exactly one ground truth generative factor present in the data \cite{kim2018disentangling}. Existing frameworks for both learning and evaluating disentangled representations \cite{higgins2017beta, kim2018disentangling, eastwood2018framework, chen2018isolating,kumar2017variational} rely on the assumption that the ground truth factors of variation are known a priori and are independent. The core idea is to measure how changes in the generative factors affect the latent representations and vice versa. In cases when an encoder network is available, this is typically achieved with a classifier that was trained to predict which generative factor in the input data was held constant given a latent representation \cite{higgins2017beta, kim2018disentangling, eastwood2018framework, kumar2017variational, chen2018isolating}. In generative models without an encoder network, such as GANs, disentanglement is measured by visually inspecting the latent traversals provided that the input data are images \cite{chen2016infogan, jeon2019ibgan, lee2020high, liu2019oogan}. However, these measures are difficult to apply when generative factors of variation are unknown or when manual visual inspection is not possible,
both of which is the case with sequences of motor commands for controlling a robotic arm. We therefore define a measure of disentanglement that does not rely on any of these requirements. Our measure, \textit{disentanglement with precision and recall (DwPR)}, is derived from the precision and recall measure introduced by \cite{kynkaanniemi2019improved} and estimates the level of disentanglement by measuring the diversity of trajectories generated when holding one latent dimension fixed.  
In contrast to existing measures, DwPR measures how changes in the latent space affect the generated trajectories in a fully unsupervised way. 
\section{Overview of the GenRL Framework}
\label{sec:preliminaries}

In this section, we provide an overview of the proposed GenRL approach. We consider a finite-horizon Markov decision process of length $T$, defined by a tuple $(\mathbf{S}, \mathbf{U}, p(s_{t+1}|s_t, u_t)$, $p(s_0), r(s_{T}))$, where $\mathbf{S}$ is a set of states $s$, $\mathbf{U}$ is a set of $M$-dimensional motor actions $u$, 
$p(s_{t+1}|s_t, u_t)$ is the state transition probability, $p(s_0)$ is the initial state distribution, 
and $r(s_{T})$ is the terminal reward function. 
The state contains information about the current configuration of the environment as well as the goal to reach in case a goal-conditioned policy has to be obtained.
By abuse of notation, we will drop the explicit distinction between a random variable, e.g., $s$, and a concrete instance of it, in the rest of the paper and only write $s$ when no confusions can arise.

We consider settings in which the agent only receives sparse terminal rewards at the end of an episode. Let $r^*$ be the minimum reward required to successfully complete the task.
We wish to find a policy $\pi_\Theta(\tau|s_0)$ which conditioned on an initial state, outputs a sequence of motor actions $\tau = \{u_{t}\}_{t=0:T-1}$ that results in high terminal rewards. 
The likelihood of a roll-out of states and actions $p_\Theta\{s_0, u_0, ..., s_{T-1}, u_{T-1}, s_T\}$ under the policy $\pi_\Theta$ is computed as  $p(s_0)$ $\prod_{t=0}^{T-1} p(s_{t+1}|s_t, u_t) \pi_\Theta (u_t | s_0)$. 
The reward can then be found as $r = r(s_T)$. 
Considering the stochasticity of the terminal state, $s_T$, the terminal reward can be modeled by a conditional distribution $p(r|s_0, \tau)$ conditioned on the initial state $s_0$, and the sequence of actions $\tau$. 
Here, the policy's objective is to maximize the likelihood of the terminal reward $r$ being greater than $r^*$
\begin{equation}
\begin{split}
    \Theta^* & = \argmax_{\Theta} \iint p(s_0)\, \pi_{\Theta}(\tau | s_0)\, p(r|s_0,\tau)\, ds \, d\tau \\
     & = \argmax_{\Theta} \mathbb{E}_{s_0 \sim p(s_0), \tau \sim \pi_{\Theta}(\tau|s_0)} [\, p(r | s_0, \tau)\,],
    \label{eq:margin_tau}
\end{split}
\end{equation}
 where, we omitted the reward threshold $r^*$ for simplicity.
Our approach is based on training a generative model $g_\vartheta$, parametrized by $\vartheta$, that maps a low-dimensional latent action sample $\alpha \in \mathbb{R}^{N_\alpha}$ into a motion trajectory $\tau \in \mathbb{R}^{T\times M}$, where $N_\alpha \ll T\times M$. 
In other words, we assume that the search space is limited to the trajectories spanned by the generative model. 
In this case, the feed-forward policy search problem splits into two sub-problems: (i) finding the mapping $g_\vartheta(\alpha)$ and (ii) finding the sub-policy $\pi_\theta (\alpha | s)$, where $\Theta = [\theta, \vartheta]$. 
Instead of marginalizing over trajectories as in~\eqref{eq:margin_tau} we marginalize over the latent variable by exploiting the generative model
\begin{equation}
\theta^* = \argmax_{\theta} \mathbb{E}_{s_0 \sim p(s_0), \alpha \sim \pi_{\theta}(\alpha|s_0)} [\, p(r|s_0, g_\vartheta(\alpha)\,)\,].
\label{eq:margin_alpha}
\end{equation}

\begin{figure}
\centering
\includegraphics[width=0.8\linewidth]{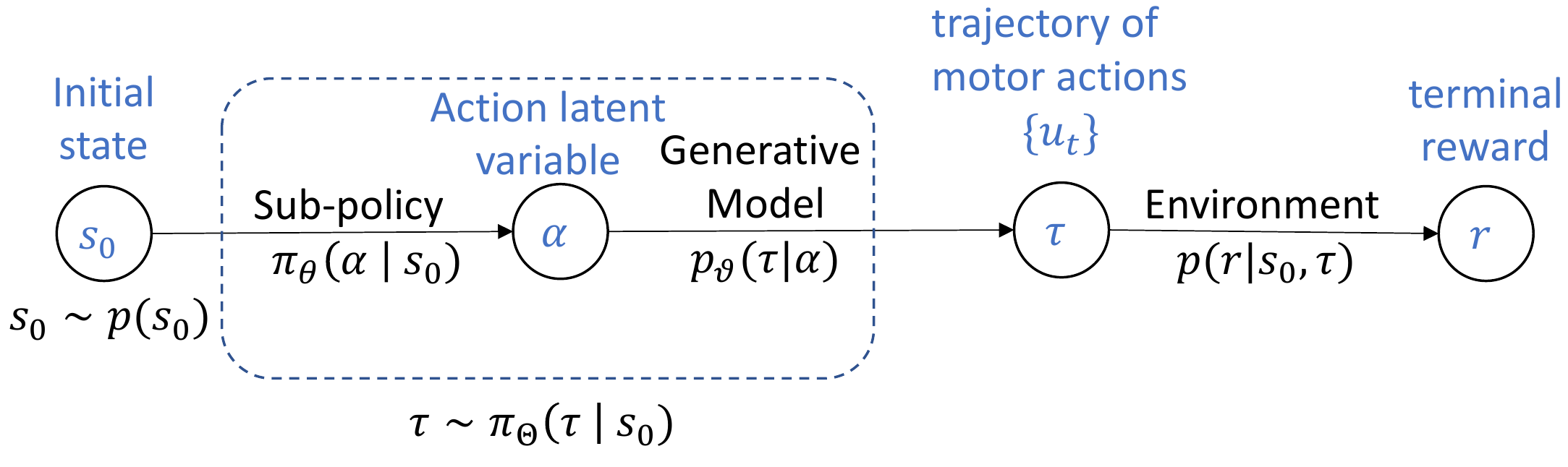} 
\caption{The architecture of the deep action-selection policy $\pi_\Theta$ based on latent-variable generative models. The policy consists of two models, the sub-policy $\pi_\theta(\alpha|s_0)$ that assigns a distribution over the action latent variable $\alpha$ conditioned on a given initial state $s_0$, and a generative model $\tau = g_\vartheta(\alpha)$ that maps a latent action sample $\alpha \sim \pi_\theta(\alpha|s_0)$  into a trajectory of motor actions $\tau$. Given the initial state $s_0$ and the action trajectory $\tau$ that is executed on the robot, the environment returns a terminal reward $r$ according to the reward probability $r \sim p(r|s_0,\tau)$.}\label{fig:deep_policy_network}
\end{figure}

An overview of our approach is shown in Figure~\ref{fig:deep_policy_network}. Once the models are trained the output of the policy $\pi_\Theta$ is found by first sampling from the sub-policy $\alpha \sim \pi_\theta(\alpha | s_0)$ given an initial state $s_0$ and then using the mapping $g_\vartheta$ to get the sequence of motor actions $\tau = g_\vartheta(\alpha)$. The state $s_0$ and the generated trajectory $\tau$ are then given to the environment which outputs a terminal reward $r$.
In the rest of the text we refer to the sub-policy as the \textit{policy} and omit the parameters $\vartheta$ from the notation $g_\vartheta$ when they are not needed.
In the following section, we introduce the expectation-maximization (EM) algorithm for training an action-selection policy using a generative model based on which different motor trajectories suitable for solving a given task can be generated. 
\section{Expectation-Maximization Policy Training}
\label{sec:em_policy_training}
The EM algorithm is a well-suited approach to find the maximum likelihood solution to the intractable marginalization over the latent variable introduced in~\eqref{eq:margin_tau} and~\eqref{eq:margin_alpha}. 
We use the EM algorithm to find an optimal policy $\pi_{\theta^*}(\alpha|s_0)$ by first introducing a variational policy $q(\alpha|s_0)$ which can be a copy of the policy network. 
In this case, the variational policy is simply the next update for the policy. 
As the goal is to find an action trajectory $\tau$ that maximizes the reward probability $p(r|s_0,\tau)$, we start by expressing its logarithm as $\log p(r | s_0) = \int q(\alpha|s_0) \log p(r|s_0) d\alpha$, where we used the identity $\int q(\alpha|s_0) d\alpha = 1$ and omitted the conditioning on $\tau$ in the reward probability for simplicity.
Following the EM derivation introduced in \cite{neumann2011variational} and using the identity $p(r|s_0) = p(r,\alpha|s_0)/p(\alpha |r,s_0)$, the expression can be further decomposed into
\begin{align}
\log p(r|s_0) & = \underbrace{\int q(\alpha|s_0) \log \frac{p(r, \alpha|s_0)}{q(\alpha|s_0)} d\alpha}_{\text{I}}  + \underbrace{\int q(\alpha|s_0) \log \frac{q(\alpha|s_0)}{p (\alpha | r, s_0)}\,d\alpha}_{\text{II}} \label{eq:marginal_decomposed} 
\end{align}

The second term (II) is the Kullback-Leibler (KL) divergence $D_{KL}( q(\alpha|s_0) \,||\, p(\alpha | r,s_0) )$ between distributions $q(\alpha|s_0)$ and $p(\alpha | r,s_0)$, which is a non-negative quantity. Therefore, the first term (I) provides a lower-bound for $\log p(r|s_0)$. 
Instead of maximizing the log-likelihood term directly, we maximize the lower bound using the EM algorithm. It is an  iterative procedure consisting of two steps known as the expectation (E-) and the maximization (M-) steps, introduced in the following sections. 

\subsection{Expectation Step } 
\label{sec:EM}
The goal of the E-step is to maximize the lower bound indirectly by minimizing the KL-divergence with respect to the variational policy. The left hand side of Eq.~\ref{eq:marginal_decomposed} does not depend on the variational policy $q(\alpha|s)$. Therefore, the sum of (I) and (II) in the equation must remain the same when we only update the variational policy. As the result, reducing (II) in Eq.~\ref{eq:marginal_decomposed} should increase (I). 
Therefore, the E-Step objective, is to minimize (II) by optimizing $q$. Assuming that $q$ is parametrized by $\phi$, the E-step objective function is given by
\begin{align}
    \phi^* &=\argmin_{\phi} D_{KL}(\,q_\phi(\alpha|s)\,||\,p(\alpha|r,s)\,) \nonumber \\
     &= \argmax_{\phi} \mathbb{E}_{\alpha \sim q_\phi(\alpha|s)}[\log p(r|s, \alpha)] -  D_{KL}(\,q_\phi(\alpha|s)\,||\,\pi_\theta(\alpha|s)\,),
    \label{eq:E_loss}
\end{align}
where, we used the Bayes rule $p(\alpha|r,s) =  p(r|\alpha,s) p(\alpha|s)/p(r|s)$ and substituted $p(\alpha|s)$ by $\pi_\theta (\alpha|s)$. 
We can replace the original MDP with another one in which the action space is the latent space of the generative model. Besides, the horizon length of this new MDP is equal to one,  $T = 1$,  since the entire sequence of actions is now modeled by a single latent variable $\alpha$. 
Therefore, we solve a contextual multi-armed bandit instead of a sequential decision making problem. The reward function can now be considered as $r(s, g(\alpha))$, or in short $r(s, \alpha)$. 
We can now use the policy gradient theorem to maximize the expected reward value $\mathbb{E}_{q_{\phi}(\alpha|s)}[r(s, \alpha)]$. 

$D_{KL}(\,q_\phi(\alpha|s)\,||\,\pi_\theta(\alpha|s)\,)$ acts as a trust region term forcing $q_\phi$ not to deviate too much from the policy distribution $\pi_\theta$. 
Therefore, we can apply policy search algorithms with trust region terms to optimize the objective given in~\eqref{eq:E_loss}. Following the derivations introduced in \cite{schulman2015trust}, we adopt TRPO objective for the E-step optimization
\begin{equation}
    \phi^* = \argmax_{\phi} \mathbb{E}_{s \sim p(s_0), \alpha \sim \pi_\theta(\alpha|s)} \left[\frac{q_\phi(\alpha|s)}{\pi_\theta(\alpha|s)}\,A(s, \alpha) - D_{KL}(q_\phi(\alpha|s) \, || \, \pi_\theta(\alpha | s))\right ],
    \label{eq:trpo}
\end{equation}
where, $A(s, \alpha) = r(s, \alpha) - V_\pi(s)$ is the advantage function, $V_\pi(s) = \mathbb{E}_{ \alpha \sim \pi_\theta(\alpha|s)} [r(s, \alpha)]$ is the value function and $\phi^*$ denotes the optimal solution for the given iteration. Note that the action latent variable $\alpha$ is always sampled from the policy $\pi_\theta(\alpha|s)$ and not from the variational policy $q_\phi(\alpha|s)$.

\subsection{Maximization Step }
The M-step yields a supervised learning objective with which we train the policy $\pi_\theta$. It directly maximizes the lower bound (I) in~\eqref{eq:marginal_decomposed} by optimizing the policy parameters $\theta$ while holding the variational policy $q_\phi$ constant. Following \cite{deisenroth2013survey} and noting that the dynamics of the system $p(r|\alpha, s)$ are not affected by the choice of the policy parameters $\theta$, we maximize (I) by minimizing the following KL-divergence
\begin{equation}
\theta^* = \argmin_{\theta} D_{KL}(\,q_\phi(\alpha|s)\, || \,\pi_{\theta}(\alpha|s)\,).
\label{eq:M_loss}
\end{equation}

In other words, the M-step updates the policy $\pi_{\theta}$ to match the distribution of the variational policy $q_\phi$ which was updated in the E-step. Similarly as in the E-step, $\theta^*$ denotes the optimal solution for the given iteration. 
However, for the variational policies 
with same network architecture as the policy network
(which is the case for the current paper), the M-step can be simply copying the parameters of the variational policy $\phi$ onto the policy parameters $\theta$. 

A summary of EM policy training is given in Algorithm \ref{alg:training}. In each iteration, a set of states $\{s_i\}$ is sampled from the initial state distribution $p(s)$. For each state $s_i$, a latent action sample $\alpha_i$ is sampled from the distribution given by the policy $\pi_\theta(\alpha|s_i)$. A generative model $g$ is then used to map every latent action variable $\alpha_i$ into a full motor trajectory $\tau_i$ which is then deployed on the robot to get the corresponding reward value $r_i$. 
In the inner loop, the variational policy $q_\phi$ and the main policy $\pi_\theta$ are updated on batches of data. 
\begin{algorithm}[h]
\SetKwInOut{Input}{Input}
\SetKwInOut{Output}{Output}
\Input{generative model $g_\vartheta$, initial policy $\pi_\theta$, initial value function $V_\pi$, \\batch size $N$}
\Output{trained $\pi_\theta$}
\While{training $\pi_\theta$}{
\For{$i = 1, \dots, N$}{
sample states $s_i \sim p(s_0)$ \\
sample actions $\alpha_i \sim \pi_\theta(.|s_i)$ \\
generate motor actions $\tau_i \leftarrow g_\vartheta(\alpha_i)$  \\
obtain the reward $r_i\leftarrow r(s_i, \tau_i)$ \\
}
\Repeat{training done}{
\textbf{E-step:} \\
\quad update the variational policy $q_\phi$ according to~\eqref{eq:trpo} given $\{s_i, \alpha_i, r_i\}_{i=1}^N$ \\
\textbf{M-step:} \\
\quad update the policy $\pi_\theta$ by copying the variational policy, i.e.  $\theta \leftarrow \phi $\\
}
update the value function $V = \argmin_{V'} \sum_{i=1}^N (V'(s_i) - r_i)^2$ given $\{s_i,r_i\}_{i=1}^N$\\
}
\vspace{0.5cm}
\label{alg:training}
\caption{GenRL framework.}
\end{algorithm}

\section{Generative Models in GenRL and their Evaluation}
\label{sec:generative_model_training}
So far we discussed how to train an action-selection policy based on the EM algorithm to regulate the action latent variable which is the input to a generative model. 
In this section, we review two prominent generative models, Variational Autoencoder (VAE) and Generative Adversarial Network (GAN), which we use in GenRL to generate sequences of actions required to solve the sequential decision-making problem. 
We then introduce a set of measures used to predict which properties of a generative model will influence the performance of the policy training. 

\subsection{Generative Models}
We aim to model the distribution $p(\tau)$ of the motor actions  that are suitable to complete a given task. To this end, we introduce a low-dimensional random variable $\alpha$ with a probability density function $p(\alpha)$ representing the latent actions which are mapped into unique trajectories $\tau$ by a generative model $g$. The model $g$ is trained to maximize the likelihood $\mathbb{E}_{\tau \sim \mathcal{D}, \alpha \sim p(\alpha)}[p_\vartheta(\tau|\alpha)]$ of the training trajectories $\tau \in \mathcal{D}$ under the entire latent variable space.

\subsubsection{Variational autoencoders}
A VAE \cite{kingma2014auto, rezende2014stochasticvae2} consists of encoder and decoder neural networks representing the parameters of the approximate posterior distribution $q_\varphi(\alpha | \tau)$ and the likelihood function $p_\vartheta(\tau|\alpha)$, respectively. The encoder and decoder neural networks, parametrized by $\varphi$ and $\vartheta$, respectively, are jointly trained to optimize the variational lower bound
\begin{equation}
     \max_{\varphi, \vartheta} \mathbb{E}_{\alpha \sim q_\varphi(\alpha|\tau)}[\log p_\vartheta(\tau|\alpha)] - \beta D_{KL}(q_\varphi (\alpha | \tau) || p(\alpha)),
\label{eq:vae}
\end{equation}
where the prior $p(\alpha)$ is the standard normal distribution and the parameter $\beta$ \cite{higgins2017beta} a variable controlling the trade-off between the reconstruction fidelity and the structure of the latent space regulated by the KL divergence. A $\beta > 1$ encourages the model to learn more disentangled latent representations \cite{higgins2017beta}.

\subsubsection{Generative adversarial networks}
A GAN model \cite{goodfellow2014generative} consists of a generator and discriminator neural network that are trained by playing a min-max game. The generative model $g_\vartheta$, parametrized by $\vartheta$, transforms a latent sample $\alpha$ sampled from the prior noise distribution $p(\alpha)$ into a trajectory $\tau = g_\vartheta(\alpha)$. The model needs to produce realistic samples resembling those obtained from the training data distribution $p(\tau)$. It is trained by playing an adversarial game against the discriminator network $D_\varphi$, parametrized by $\varphi$, which needs to distinguish a generated sample from a real one. The competition between the two networks is expressed as the following min-max objective
\begin{align}
    \min_\vartheta \max_\varphi \mathbb{E}_{\tau \sim p(\tau)} [\log D_\varphi(\tau)] + \mathbb{E}_{\alpha \sim p(\alpha)} [\log(1 - D_\varphi(g_\vartheta(\alpha))]. \label{eq:gan_original}
\end{align}
However, the original GAN formulation~\eqref{eq:gan_original} does not impose any restrictions on the latent variable $\alpha$ and therefore the generator $g_\vartheta$ can use $\alpha$ in an arbitrary way. To learn disentangled latent representations we instead use InfoGAN \cite{chen2016infogan} which is a version of GAN with an information-theoretic regularization added to the original objective. The regularization is based on the idea to maximise the mutual information $I(\alpha, g_\vartheta(\alpha))$ between the latent code $\alpha$ and the corresponding generated sample $g_\vartheta(\alpha)$. An InfoGAN model is trained using the following information minmax objective \cite{chen2016infogan}
\begin{align}
    \min_{\vartheta, \psi} \max_\varphi \, & \mathbb{E}_{\tau \sim p(\tau)} [\log D_\varphi(\tau)] + \mathbb{E}_{\alpha \sim  p(\alpha)} [\log(1 - D_\varphi(g_\vartheta(\alpha))] \nonumber \\ 
    &- \lambda \mathbb{E}_{\alpha \sim p(\alpha), \tau \sim g_\vartheta(\alpha)}[\log Q_\psi(\alpha | \tau)], \label{eq:gan}
\end{align}
where $Q_\psi(\alpha | \tau)$ is an approximation of the true unknown posterior distribution $p(\alpha | \tau)$ and $\lambda$ a hyperparameter. In practice, $Q_\psi$ is a neural network that models the parameters of a Gaussian distribution and shares all the convolutional layers with the discriminator network $D_\varphi$ except for the last few output layers.

\subsection{Evaluation of Generative Models in GenRL}
\label{sec:eval_generative_model}
We review the characteristics of generative models that may lead to effective policy training by measuring precision and recall as well as disentanglement. Our goal is to be able to judge the quality of the policy training by evaluating the generative models prior to the RL training. We relate the measures to the performance of the policy training in Section~\ref{sec:exp:eval_gen_models}.

\subsubsection{Precision and recall} \label{sec:prec_and_recall}
A generalized precision and recall measure for comparing distributions was introduced by \cite{sajjadi2018assessing}, and further improved by \cite{kynkaanniemi2019improved}, to evaluate the quality of a distribution learned by a generative model $g$.
It is based on the comparison of samples obtained from $g$ with the samples from the ground truth reference distribution. In our case, the reference samples correspond to the training motor trajectories. Intuitively, \textit{precision} measures the quality of the generated sequences of motor actions by quantifying how similar they are to the training trajectories. It determines the fraction of the generated samples that are realistic. On the other hand, \textit{recall} evaluates how well the learned distribution covers the reference distribution and determines the fraction of the training trajectories that can be generated by the generative model. In the context of the policy training, we would like the output of $\pi_\Theta$ to be as similar as possible to the demonstrated motor trajectories. It is also important that $\pi_\Theta$ covers the entire state space as it must be able to reach different goal states from different task configurations. Therefore, the generative model should have both high precision and high recall scores. 

The improved measure introduced by \cite{kynkaanniemi2019improved} is based on an approximation of the
manifolds underlying the training and generated data. In particular, given a set $\boldsymbol{T} \in \{\boldsymbol{T_r}, \boldsymbol{T_g}\}$ of either real training trajectories $\boldsymbol{T_r}$ or generated trajectories $\boldsymbol{T_g}$, the corresponding manifold is estimated by forming hyperspheres around each trajectory $\tau \in \boldsymbol{T}$ with radius equal to its $k$th nearest neighbour $\NN_k(\tau, \boldsymbol{T})$. To determine whether or not a given novel trajectory $\tau$ lies within the volume of the approximated manifold we define a binary function
\[
 f(\tau, \boldsymbol{T}) =
  \begin{cases}
  1 & \text{if $||\tau - \tau||_2 \le ||\tau - \NN_k(\tau, \boldsymbol{T})||_2$ for at least one $\tau \in \boldsymbol{T}$} \\
  0 & \text{otherwise.} 
  \end{cases}
\]
By counting the number of generated trajectories $\tau_g \in \boldsymbol{T}_g$ that lie on the manifold of the real data $\boldsymbol{T}_r$ we obtain the \textit{precision}, and similarly the \textit{recall} by counting the number of real trajectories $\tau_r \in \boldsymbol{T}_r$ that lie on the manifold of the generated data $\boldsymbol{T}_g$
\begin{align}
    \text{precision}(\boldsymbol{T}_r, \boldsymbol{T}_g) = \frac{1}{|\boldsymbol{T}_g|} \sum_{\tau_g \in \boldsymbol{T}_g} f(\tau_g, \boldsymbol{T}_r) \quad \text{and} \quad
    \text{recall}(\boldsymbol{T}_r, \boldsymbol{T}_g) = \frac{1}{|\boldsymbol{T}_r|} \sum_{\tau_r \in \boldsymbol{T}_r} f(\tau_r, \boldsymbol{T}_g). \label{eq:prec_and_recall}
\end{align}
In our experiments, we use the original implementation provided by \cite{kynkaanniemi2019improved} directly on the trajectories as opposed to their representations as suggested in their paper.

\subsubsection{Disentanglement with precision and recall} \label{sec:disentaglement_with_prec_and_recall}
In addition to evaluating the quality of the generated samples, we also assess generative models by the quality of their latent action representations in terms of disentanglement. We evaluate it with a novel measure, called \textit{disentanglement with precision and recall (DwPR)}, that we derive from the precision and recall metric introduced in Section~\ref{sec:prec_and_recall} as described below.

A disentangled representation of the motor data obtained from the latent space of a generative model can be defined as the one in which every end state of the system is controllable by one latent dimension determined by the vector basis of the latent space. For example, consider a task where the goal is to shoot a hockey puck to a specific position on a table top. We say that a latent representation given by a generative model $g$ is well-disentangled if there exists a basis of the latent space with respect to which each dimension controls one axis of the end position of the hockey puck. Our hypothesis is that the more disentangled the representation is, the more efficient is the policy training. 
We experimentally evaluate the effect of disentangled representations on the performance of the policy training in Section~\ref{sec:exp:eval_gen_models}.

Our measure of disentanglement is based on the following observation: if representations given by $g$ are well disentangled, then setting one latent dimension to a fixed value should result in limited variation in the corresponding generated trajectories $\boldsymbol{T_g}$. For example, if the $1$st latent dimension controls the distance of the end position of the hockey puck to a fixed origin then setting it to a fixed value should limit the set of possible end positions of the puck to a certain spherical range (assuming the polar coordinates in the end state space). In this case, assuming that $g$ well reflects the true data distribution, we should obtain a high precision score since the limited end positions are close to the true ones but a low recall since these do not cover all possibilities.
We present the details below.

Our approach can be described in two phases. In phase 1, we generate the two sets of generated trajectories on which we run the precision and recall metrics from Section~\ref{sec:prec_and_recall}. Firstly, for a fixed latent dimension $l \in \{1, \dots, N_\alpha\}$ we perform a series of $D \in \mathbb{N}$ \textit{latent interventions} where we set the $l$th component of a latent code $\alpha_l$ to a fixed value $I_d$, $\alpha_l = I_d$ for $d = 1, \dots, D$. Each intervention $\alpha_l = I_d$ is performed on a set of $n$ samples sampled from the prior distribution $p(\alpha)$. We denote by $\boldsymbol{T}_g^{l-I_d}$ the set of $n$ generated trajectories corresponding to the latent samples on which we performed the intervention $\alpha_l = I_d$. We define $\boldsymbol{T}_g^{l} = \cup_{d = 1}^D \boldsymbol{T}_g^{l-I_d}$ to be the union of the trajectories corresponding to all $D$ interventions performed on the fixed latent dimension $l$. By construction,  $\boldsymbol{T}_g^{l}$ consists of $n \cdot  D$ trajectories. For example, $D = 5$ latent interventions on the $2$-st latent dimension yield the set $\boldsymbol{T}_g^{2} = \boldsymbol{T}_g^{2-I_1} \cup \cdots \cup \boldsymbol{T}_g^{2-I_5}$. Secondly, we define a set $\boldsymbol{T}_g^{p(\alpha)}$ of $n\cdot D$ generated trajectories corresponding to randomly sampled latent samples from the prior distribution $p(\alpha)$. We then treat $\boldsymbol{T}_g^{p(\alpha)}$ and $\boldsymbol{T}_g^{l}$ as $\boldsymbol{T}_r$ and $\boldsymbol{T}_g$ in~\eqref{eq:prec_and_recall}, respectively, and calculate $\text{precision}(\boldsymbol{T}_g^{p(\alpha)}, \boldsymbol{T}_g^{l})$ and $\text{recall}(\boldsymbol{T}_g^{p(\alpha)}, \boldsymbol{T}_g^{l})$ for every latent dimension $l$. Note that we consider $\boldsymbol{T}_g^{p(\alpha)}$ as the reference set $\boldsymbol{T}_r$ because we wish to evaluate the disentanglement relative to the generative capabilities of each individual model $g$.  

In phase 2, our goal is to aggregate the  precision and recall values obtained in phase 1 to measure the magnitude of the disentanglement of the latent dimensions. To this end, we define
\begin{align}
\delta(l) = \text{precision}(\boldsymbol{T}_g^{p(\alpha)}, \boldsymbol{T}_g^{l}) - \text{recall}(\boldsymbol{T}_g^{p(\alpha)}, \boldsymbol{T}_g^{l}) \label{eq:disentanglement}
\end{align}
for each latent dimension $l$.
Since a well disentangled generative model $g$ should achieve high precision but low recall on $\boldsymbol{T}_g^{p(\alpha)}$ and $\boldsymbol{T}_g^{l}$, the higher the $\delta(l)$ value, the higher the magnitude of disentanglement for the latent dimension $l$. Note that this definition correctly reflects even cases where the trajectories $\boldsymbol{T}_g^{l}$ exhibit very low variance as this will be reflected in a low precision, thus yielding a low $\delta(l)$. Due to the nature of the robotics tasks considered in our experiments in Section~\ref{sec:experiments}, where the end state spaces are two dimensional, we compare our models based on the highest two $\delta(l)$ values obtained for each $g$. 
Let $\delta_1(g)$ and $\delta_2(g)$ denote the highest and second highest $\delta(l)$ values, respectively. 
To compare the magnitude of the two most disentangled latent dimensions among the models, we use $\delta_1(g)$ and the average $\widetilde{\delta}(g) := \frac{\delta_1(g) + \delta_2(g)}{2}$.
Note that we generally cannot determine whether the two latent dimensions control the same aspect of the end state space without executing the obtained trajectories on the robot.

\section{Experiments}
\label{sec:experiments}
We experimentally validated our approach by:
\begin{enumerate}
    \item evaluating data efficiency of GenRL (Section~\ref{sec:exp:data-efficiency}), 
    \item evaluating safety of GenRL (Section~\ref{sec:exp:safety}), and
    \item determining the characteristics of generative models that contribute to a more data-efficient policy training (Section~\ref{sec:exp:eval_gen_models}).
\end{enumerate}
We performed several experiments on two simulated tasks: \textit{hockey task} and \textit{basketball task}, described in Section~\ref{sec:experimental_setup:tasks} and compared the performance of the GenRL framework to two state-of-the-art policy search algorithms, PPO \cite{schulman2017proximal} and SAC \cite{haarnoja2018soft}.

\subsection{Experimental Setup and Implementation Details} \label{sec:exp:details}
In this section, we provide details of the experimental setup. We describe the considered hockey and basketball tasks in Section~\ref{sec:experimental_setup:tasks}, provide details on the training and evaluation of the generative models in Section~\ref{sec:experimental_setup:generative_model} and the corresponding EM policy training in Section~\ref{sec:experimental_setup:policy}. Lastly, we describe the RL benchmark methods used as baselines in Section~\ref{sec:experimental_setup:RL_benchmarks}.

\subsubsection{Evaluation tasks} \label{sec:experimental_setup:tasks}
\begin{figure}
\centering
\includegraphics[width=0.8\linewidth]{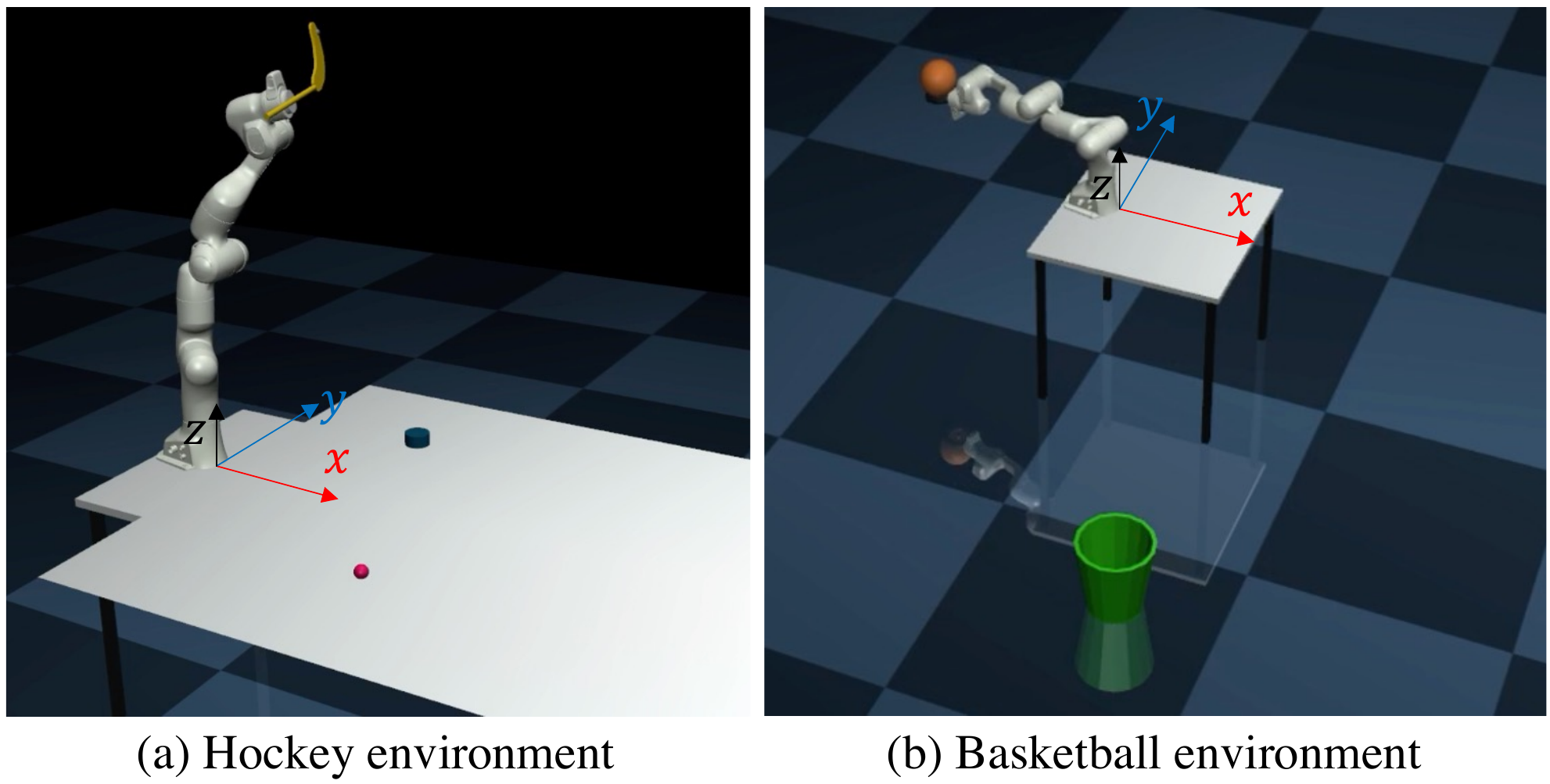} 
\caption{The hockey shooting and basketball throwing environments used in the experiments. The Mujoco engine is used to simulate the robotic environments. }
\label{fig:envs}
\end{figure}

As illustrated in Figure \ref{fig:envs}, we evaluated GenRL by training goal-conditioned policies for two robotic tasks, shooting a hockey puck and throwing a basketball.

\textbf{Hockey} In this task, the goal is to learn to shoot a hockey puck to a target position (illustrated by the red dot in Figure~\ref{fig:envs}) by a hockey stick which is attached to the robot end-effector. 
The target is randomly placed at a position $(x, y)$ in a rectangular area with $x \in [0.65, 1.3]$ meters, and $y \in [-0.4, 0.4]$ meters from the robot. 
The feed-forward policy, conditioned on the $xy$ coordinates of the goal position, generates a sequence of actions consisting of 7 motor actions for 69 time-steps. The \textit{feedback} policies, conditioned on the $xy$ coordinate of the goal position (2-dimensional) and the joint position (7-dimensional) and velocity of the robot (7-dimensional), assign a distribution over the next motor action (7-dimensional) which is the commanded change in the joint position of the robot. 
The agent receives a terminal reward at the end of each episode which is calculated as the negative Euclidean distance between final position of the puck and the target. 

\textbf{Basketball} In this task, the goal of the robot is to throw a ball to a basket which is randomly placed at a radius of one to eight meters from the robot, and at the angle formed by the basket center and the x-axis ranging from $-\pi/2$ to $\pi/2$.  
The feed-forward policy, conditioned on the $xy$ coordinates of the basket, generates a sequence of actions consisting of 7 motor actions for 78 time-steps. On the other hand, the \textit{feedback} policies, which are conditioned on the $xy$ coordinate of the basket and the joint position and velocity of the robot (14 dimensions), assign a distribution over the next motor action (7 dimensions). 
In this environment, the reward is calculated as follows:
at each timestep where the ball passes the height of the basket, the Euclidean distance to the goal in the horizontal plane is calculated and stored.
At the end of the episode, the minimum of these distances is used as the terminal reward (with a minus sign).
This allows the agent to learn complex throwing strategies (such as ones where the ball bounces on the floor multiple times before landing in the basket) without making the reward excessively sparse.

\subsubsection{Generative model training and evaluation}
\label{sec:experimental_setup:generative_model}
The generative models are represented by neural networks that map a low-dimensional action latent variable $\alpha$ into a $n_{\text{joints}} \times n_{\text{time-steps}}$ dimensional vector representing $n_{\text{joints}}$ motor actions and $n_{\text{time-steps}}$ time-steps. For all tasks $n_{\text{joints}} = 7$, while $n_{\text{time-steps}} = 69$ for the hockey task and $n_{\text{time-steps}} = 78$ for the basketball task. For all the models, we varied the dimension of the latent space $N_\alpha \in \{2, 3, 6\}$. For each task and each dimension $N_\alpha$, we trained \textit{i)} $3$ $\beta$-VAE models with $\beta \in 0.01, 0.005, 0.001$ which are referred to as \textit{VAE1-9} as reported in Table~\ref{tab:models} (left column), and \textit{ii)} $3$ InfoGAN models with $\gamma \in 0.1, 1.5, 3.5$ which are referred to as \textit{GAN1-9} (right column). In total, we considered $9$ $\beta$-VAE models and $9$ InfoGAN models per task. 
We refer the reader to Appendix \ref{app:gen_models_details} for the complete training details and the exact architecture of the models, as well as to Appendix~\ref{app:gen_models_traj} for visualized examples of generated trajectories. The prior distribution $p(\alpha)$ was chosen to be the standard normal $\text{N}(0, 1)$ in case of $\beta$-VAEs, and uniform $\text{U}(-1, 1)$ in case of InfoGANs. We evaluated all the generative models using the precision and recall as well as disentanglement measures introduced in Section~\ref{sec:eval_generative_model} with hyperparameters described below. 

\begin{table}
    \centering
    \vspace{0.2cm}
    
    \begin{tabular}{r|c|l|c|c||r|c|c|c|c}
     & \multirow{2}{*}{$N_\alpha$} & \multirow{2}{*}{$\beta$} & \multicolumn{2}{c||}{reward} &      & \multirow{2}{*}{$N_\alpha$} & \multirow{2}{*}{$\lambda$} & \multicolumn{2}{c}{reward} \\
     &                    &                       & \textit{hockey}     & \textit{basketball}  &      &  & & \textit{hockey}     & \textit{basketball}    \\
     \hline
VAE1 & 2                  & $0.01$                & -0.267  & -0.106 & GAN1 & 2  & $0.1$                   & -0.144 & -0.767              \\
\textbf{VAE2} & 2                  & $0.005$               & -0.085  & \cellcolor[gray]{0.65}-0.081 & GAN2 & 2  & $1.5$                   & -0.129 & -0.469               \\
\textbf{VAE3} & 2                  & $0.001$               & \cellcolor[gray]{0.85}-0.040  & \cellcolor[gray]{0.65}-0.081 & \textbf{GAN3} & 2  & $3.5$                   & \cellcolor[gray]{0.85}-0.119 & -1.183              \\
VAE4 & 3                  & $0.01$                & -0.266  & -0.109 & GAN4 & 3  & $0.1$                   & -0.155 & -0.413              \\
VAE5 & 3                  & $0.005$               & -0.060  & -0.119 & \textbf{GAN5} & 3  & $1.5$                   & \cellcolor[gray]{0.65}-0.068 & \cellcolor[gray]{0.85}-0.357              \\
\textbf{VAE6} & 3                  & $0.001$               & \cellcolor[gray]{0.65}-0.033  & -0.093 & \textbf{GAN6} & 3  & $3.5$                   & \cellcolor[gray]{0.75}-0.101 & -1.019               \\
\textbf{VAE7} & 6                  & $0.01$                & -0.266  & \cellcolor[gray]{0.85}-0.088 & \textbf{GAN7} & 6  & $0.1$                   & -0.159 & \cellcolor[gray]{0.75}-0.337              \\
VAE8 & 6                  & $0.005$               & -0.065  & -0.108 & \textbf{GAN8} & 6  & $1.5$                   & -0.153 & \cellcolor[gray]{0.65}-0.254              \\
\textbf{VAE9} & 6                  & $0.001$               & \cellcolor[gray]{0.75} -0.038 & -0.104 & GAN9 & 6  & $3.5$                   & -0.135 & -1.344              
\end{tabular}
\vspace{0.5cm}
    \caption{Training details of the generative models used in our experiments. We varied the latent dimensionality $N_\alpha$ and hyperparameters $\beta$ in case of $\beta$-VAEs and $\lambda$ in case of InfoGANs affecting the disentanglement. We additionally report the maximum reward achieved by each of these models over $3$ independent runs (see Section~\ref{sec:experimental_setup:policy}). We marked three best performing models per task in gray. } 
    \label{tab:models}
\end{table}

 \textbf{Precision and recall}
For each generative model $g$, we randomly sampled $15000$ samples from the latent prior distribution $p(\alpha)$. The corresponding set of the generated trajectories $\boldsymbol{T_g}$ was compared to a set $\boldsymbol{T_r}$ comprised of $15000$ randomly chosen training trajectories  which were sampled only once and fixed for all the models. The neighbourhood size $k$ was set to $3$ as suggested by \cite{kynkaanniemi2019improved}.

 \textbf{Disentanglement} 
For each generative model $g$, we performed $D = 5$ interventions on every latent dimension $l \in \{1, \dots, N_\alpha\}$ and sampled $n = 10000$ latent vectors for each $D$. Thus, the resulting sets $\boldsymbol{T}_g^l$ contained $n\cdot D = 50000$. We additionally generated $50000$ trajectories from the corresponding prior $p(\alpha)$ used as the reference set $\boldsymbol{T}_g^{p(\alpha)}$, which we kept fixed for each $g$. As in precision and recall, we used the neighbourhood size $k = 3$.

\subsubsection{EM Policy training} \label{sec:experimental_setup:policy}
\begin{figure}[t]
\centering
\includegraphics[width=1.0\linewidth]{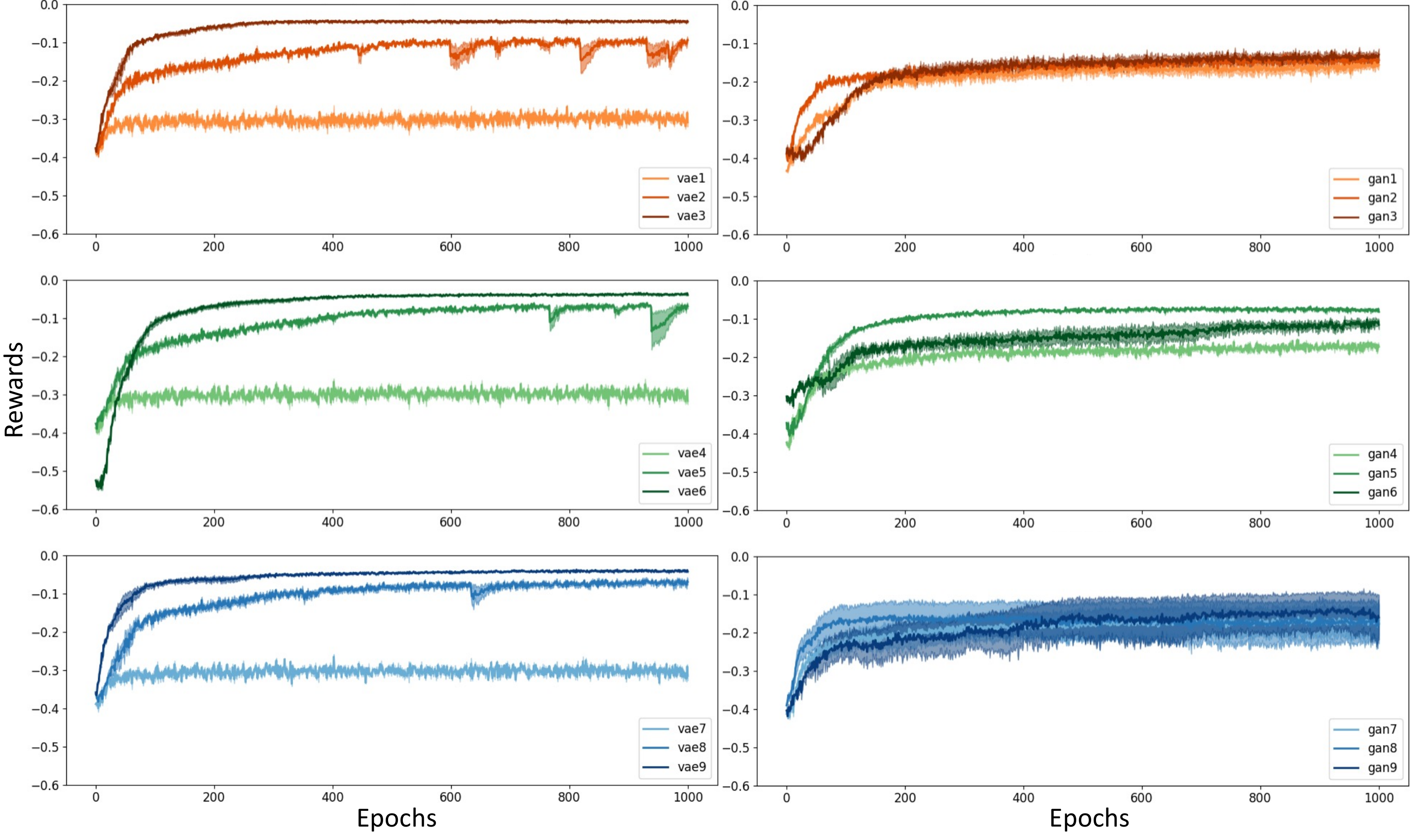}
\caption{Policy training performance for the hockey task. We report the average reward together with the standard deviation  obtained during EM policy training across three different random seeds. }
\label{fig:hockey_results}
\end{figure}

\begin{figure}[h]
\centering
\includegraphics[width=1.0\linewidth]{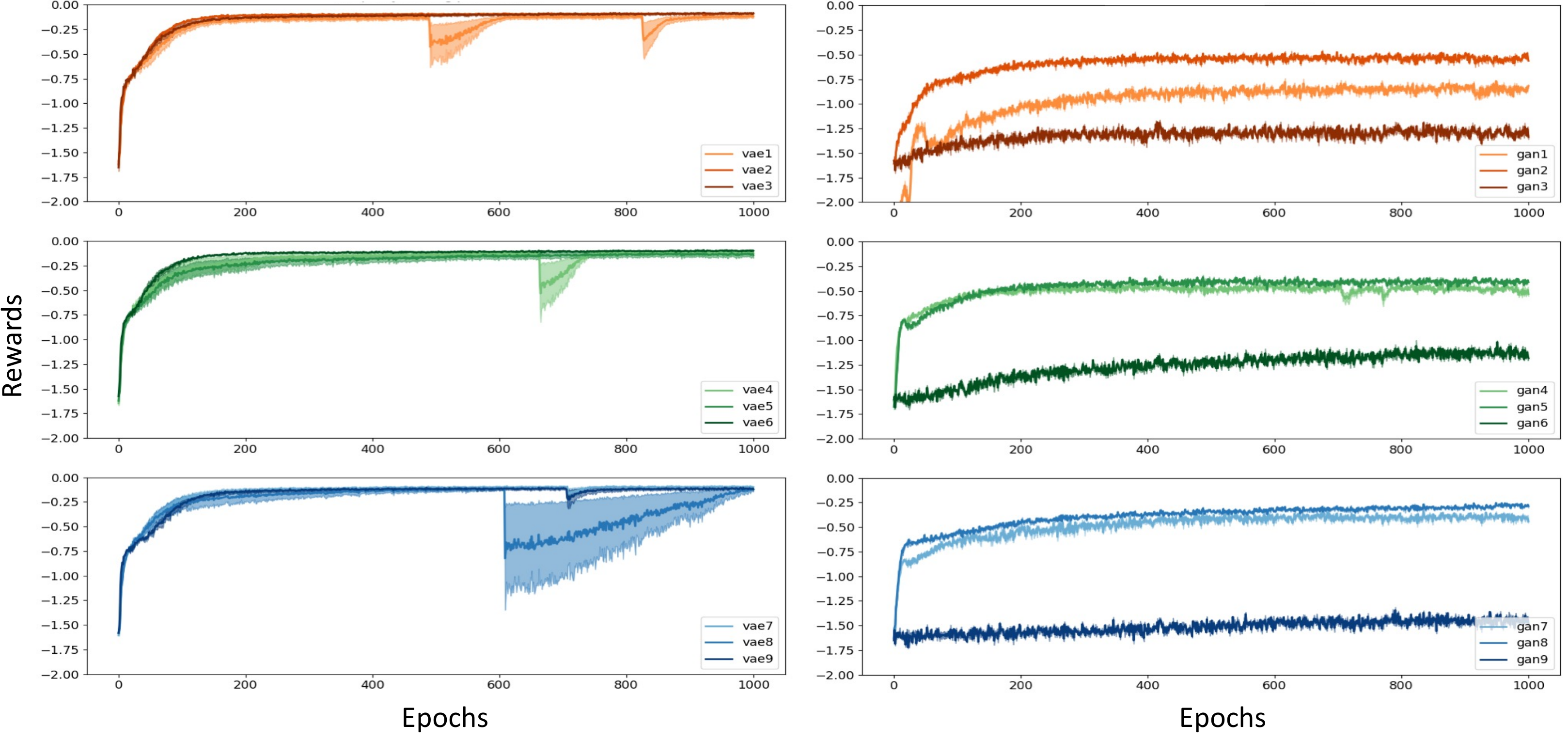}
\caption{Policy training performance for the basketball task. We report the average reward together with the standard deviation obtained during EM policy training  across three different random seeds. }
\label{fig:hbasketball_results}
\end{figure}

For each generative model and each task, we trained one policy with three different random seeds. The obtained average training performances together with the standard deviations are shown in Figures~\ref{fig:hockey_results} and~\ref{fig:hbasketball_results} for the hockey and basketball tasks, respectively. Using these results, we labeled each generative model with the
maximum reward achieved during the EM policy training across all three random seeds. The values are reported Table~\ref{tab:models} where we marked the best performing models per task in gray. In general, we observe that while VAE3 and GAN6 performed well on both tasks, VAEs generally performed better than GAN models.

\subsubsection{RL benchmarks} 
\label{sec:experimental_setup:RL_benchmarks}
We compared GenRL with two state-of-the-art policy search algorithms, PPO \cite{schulman2017proximal} and SAC \cite{haarnoja2018soft}, in terms of the data efficiency and safety of RL exploration. 
GenRL trained a generative model with offline action data containing 40K episodes of 7-dimensional motor actions, each containing 69 time-steps for the hockey task and 78 time-steps for the basketball task.
We used the combination of human demonstration and expert trajectory shaping to collect the motor action data. First, we collected a few human demonstrations through kinesthetic teaching and then added random values to some of the robot joint velocities to generate different hits or throws, e.g., with different strengths or different orientations.
Therefore, to make fair comparisons, we also pre-trained PPO and SAC with the same amount of data. However, for the PPO and SAC methods, we needed to provide corresponding state transition and reward data in addition to the action data. We pre-trained a goal-conditioned PPO policy with the behavior cloning approach using the pre-recorded state-action data. 
On the other hand, SAC was pre-trained by adding the state-action data to the SAC's replay buffer. In this way, we ensured that PPO and SAC receive the same amount of data for pre-training the policy.

\begin{figure}[h]
\centering
\includegraphics[width=0.9\linewidth]{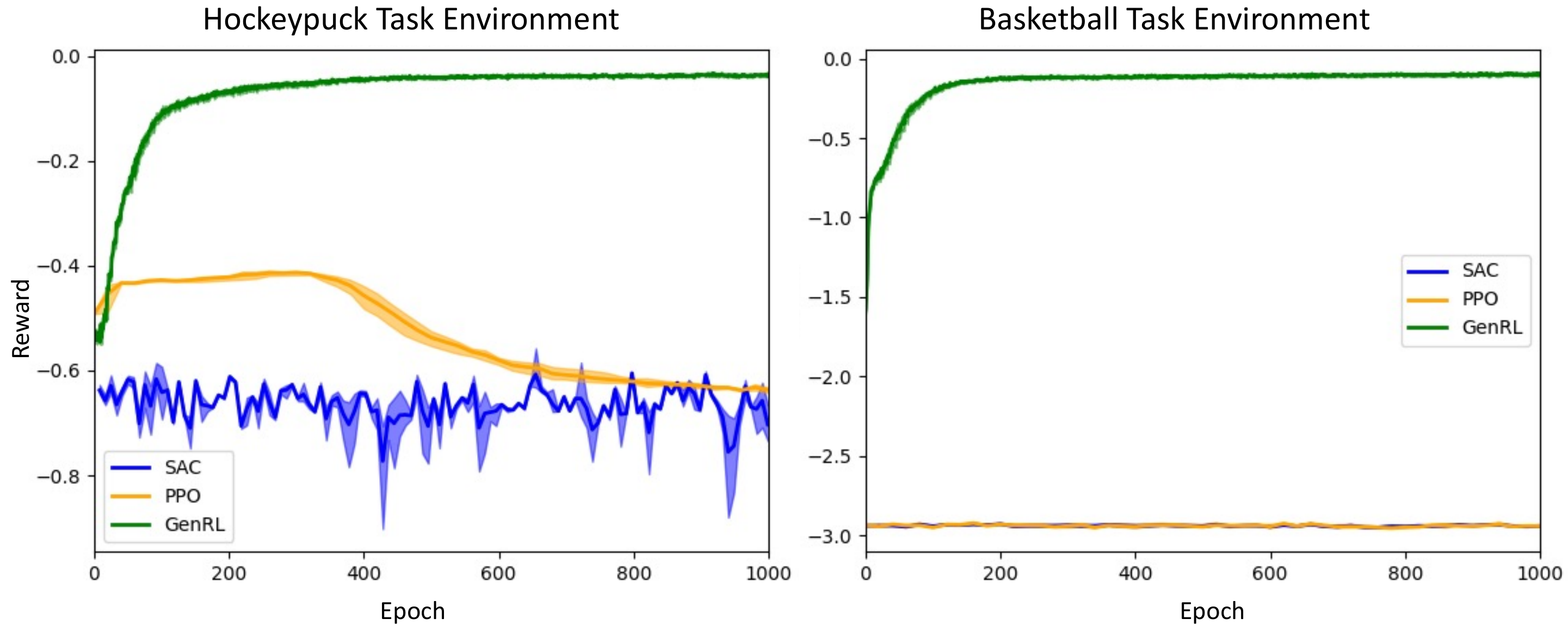}
\caption{The average policy training performance of GenRL, PPO and SAC on the hockey and basketball tasks. We report the average reward together with the standard deviation ($y$-axis) obtained across three different random seeds.  }
\label{fig:ppo_sac_performance}
\end{figure}

\subsection{Comparison to Other Policy Search Methods} 
\label{sec:exp:data-efficiency}
In this section, we compare the policy training performance of GenRL, PPO and SAC on the robotics tasks introduced in Figure~\ref{fig:envs}. 
As explained in Section \ref{sec:experimental_setup:RL_benchmarks}, we pre-train SAC and PPO with the same number of samples used to train the generative model of the GenRL method.  
A goal-conditioned PPO policy is initialized through behavior cloning provided the state-action pair at every time-step and the goal state. 
The SAC policy is initialized by adding the goal state, state-action transitions and the rewards at every time-step to the replay buffer. 
Figure~\ref{fig:ppo_sac_performance} shows the average training performance for three random seeds of SAC, PPO and GenRL for the hockey and basketball task. 
As shown, PPO and SAC cannot train a policy that can solve the tasks, and training with the pre-collected data does not help. 
For the basketball task (right), the robot looses the ball at the beginning of episodes because of the shaky motion caused by sampling actions from the PPO or SAC policies.  This is not a problem for GenRL because it uses the generative model to generate smooth trajectories of actions to properly throw the ball. 
For the hockey task (left), pre-training the policy with behavior cloning helps at the beginning of the training. However, further training with PPO worsen the learning performance most likely because it is hard to learn from the sparse rewards provided for the PPO training phase. Moreover, SAC completely fails to learn a policy for the hockey task even though its replay buffer is initialized by high-quality demonstration data. 
GenRL, on the other hand, consistently achieves significantly better performance on both of the tasks even though its generative model is only trained with action data, and not with all data containing state-action transitions and rewards. 

\begin{figure}[h]
\centering
\includegraphics[width=1.0\linewidth]{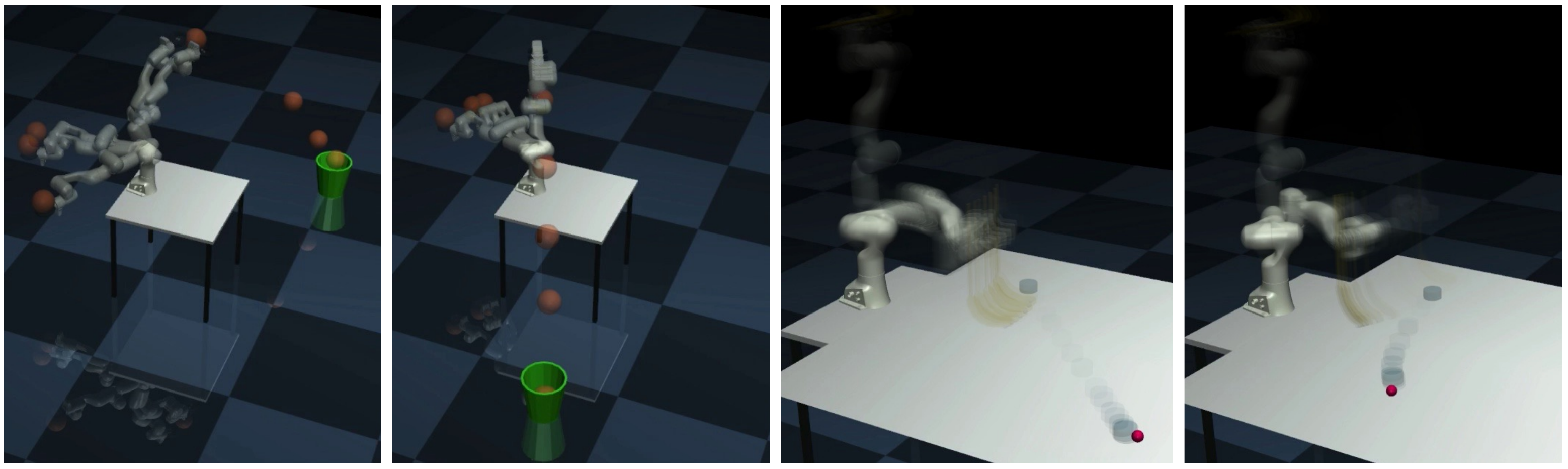}
\caption{Examples of the performance of policies trained with the GenRL method on the basketball and hockey tasks. For the basketball task, the trained policy can successfully throw the ball into the green basket placed at different random locations, and for the hockey task, the policy can shoot the puck to random target positions illustrated by the red dots. }
\label{fig:hits}
\end{figure}

Figure~\ref{fig:hits} illustrates examples of GenRL performance at the deployment phase. For the basketball task, the policy trained with GenRL can successfully throw the ball into baskets placed randomly at different locations. Also, for the hockey task, the trained policy can successfully shoot the puck to different random target positions. 

\begin{figure}[h]
\centering
\includegraphics[width=0.6\linewidth]{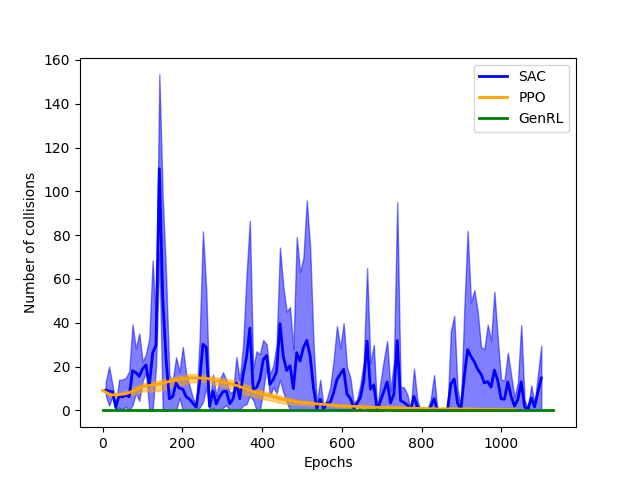}
\caption{The average number of unsafe collisions during policy training with GenRL, PPO and SAC.
We report the average number of time-steps per episode during which the hockey blade is in contact with the table top for the hockey task. }
\label{fig:collision}
\end{figure}

\subsection{Safety of Policy Training with GenRL} \label{sec:exp:safety}
In this section, we study the safety aspect of policy training on physical systems. 
Safe policy training is particularly important for real-robot applications to avoid damages to the robot and its environment due to the trial-and-error exploration of RL training. 
For this experiment, we train policies using GenRL, PPO and SAC on the hockey task, and report the average number of time-steps per episode during which the hockey blade is in contact with the table top. The collision between the blade and the table top is a proper indication of the safety of the policy training algorithm in our hockey task setting because collisions between the blade and the table top are more likely than any other type of collisions in this setting, and in real world, training must be stopped when such collisions happen. 

Similar to the experiments introduced in the previous section, we pre-train PPO and SAC with the same amount of data used to train the generative model of the GenRL. We train policies using all three methods for a fixed number of epochs. 
Figure \ref{fig:collision} illustrates the result for the GenRL, PPO and SAC methods.
Most importantly, during the entire training time with GenRL, there is no collision between the table top and the blade. According to our collision tracker algorithm,
the only collision is hitting the puck with the hockey blade. 
However, during policy training with PPO and SAC, the blade hits the table top for many time-steps which indicates that these methods may not be suitable for real-robot training. 
Therefore, considering both the performance and safety, policy training with GenRL is a better alternative than SAC and PPO when training on similar robotic tasks.

\subsection{Evaluation of Generative Models} \label{sec:exp:eval_gen_models}
In this section, we empirically determine the characteristics of generative models that contribute to a more efficient policy training for the hockey and basketball tasks. We evaluated the trained $\beta$-VAE and InfoGAN models (Section~\ref{sec:experimental_setup:generative_model}) by measuring precision and recall and disentanglement measures introduced in Section~\ref{sec:eval_generative_model}. Using these models we trained several RL policies (as described in Section~\ref{sec:experimental_setup:policy}) with the proposed GenRL framework presented in Section~\ref{sec:em_policy_training}, and investigated the relation between the properties of the generative models and the performance of the policy. \textit{Our analysis in the remained of this section empirically shows that generative models that are able to cover the distribution of the training trajectories well and additionally have well disentangled latent action representations capturing all aspects of the robotic task result in a successful policy training when combined with GenRL.}

\begin{table}[t]
    \centering
    \vspace{0.2cm}
    
    \begin{tabular}{c|r|c|c|c|c}
     \multicolumn{2}{c|}{\textit{hockey}}
    & \textbf{Precision} & \textbf{Recall} & $\boldsymbol{\delta_1}$ & $\boldsymbol{\widetilde{\delta}}$ 
    \\
    \hline
    \multirow{2}{*}{VAE}  &  Pearson's R & $-0.052$ &  \cellcolor[gray]{0.641}$0.959$ &  $-0.533$ &  \cellcolor[gray]{0.621}$0.979$ \\
     & p-value & $0.893$ &  \cellcolor[gray]{0.641}$0.000$ &  $0.140$ &  \cellcolor[gray]{0.621}$0.000$ \\
    
    \hline
    \multirow{2}{*}{GAN} & Pearson's R & $-0.485$ &  $0.526$ &  $0.505$ &  $0.393$ \\
     & p-value & $0.186$ &  $0.146$ &  $0.165$ &  $0.295$  \\
     
    \hline
    \multirow{2}{*}{ALL} & Pearson's R & $-0.068$ &  \cellcolor[gray]{0.863}$0.737$ &  $0.082$ &  \cellcolor[gray]{0.767}$0.833$  \\
    & p-value & $0.789$ &  \cellcolor[gray]{0.863}$0.000$ &  $0.746$ &  \cellcolor[gray]{0.767}$0.000$ \\

    \end{tabular}
    \vspace{0.5cm}
    \caption{Correlation results between the performance of the policy reported in Table~\ref{tab:models} and resulting evaluation measures described in Section \ref{sec:eval_generative_model} for the generative models trained on the hockey task. We report Pearson correlation coefficient R (the higher the absolute value, the better $\uparrow$) with the corresponding p-value. } 
    \label{tab:hockey_pear_coef}
\end{table} 

\begin{table}[t]
    \centering
    \vspace{0.2cm}
    
    \begin{tabular}{c|r|c|c|c|c}
    \multicolumn{2}{c|}{\textit{basketball}}
    & \textbf{Precision} & \textbf{Recall} & $\boldsymbol{\delta_1}$ & $\boldsymbol{\widetilde{\delta}}$ 
    \\
    \hline
    \multirow{2}{*}{VAE}  &  Pearson's R & \cellcolor[gray]{0.664} $-0.936$ &  $0.571$ &  $-0.059$ &  $0.131$ \\
     & p-value & \cellcolor[gray]{0.664}$0.000$ &  $0.108$ &  $0.880$ &  $0.738$ \\
    
    \hline
    \multirow{2}{*}{GAN} & Pearson's R & $-0.035$ &  \cellcolor[gray]{0.789}$0.811$ &  $0.539$ &  $0.591$ \\
     & p-value & $0.929$ &  \cellcolor[gray]{0.789}$0.008$ &  $0.134$ &  $0.094$  \\
     
    \hline
    \multirow{2}{*}{ALL} & Pearson's R & $0.434$ &  \cellcolor[gray]{0.693}$0.907$ &  \cellcolor[gray]{0.952}$0.648$ &  \cellcolor[gray]{0.971}$0.629$   \\
    & p-value &  $0.072$ &  \cellcolor[gray]{0.693}$0.000$ &  \cellcolor[gray]{0.952}$0.004$ &  \cellcolor[gray]{0.971}$0.005$ \\

    \end{tabular}
    \vspace{0.5cm}
    \caption{Correlation results between the performance of the policy reported in Table~\ref{tab:models} and resulting evaluation measures described in Section \ref{sec:eval_generative_model} for the generative models trained on the basketball task. We report Pearson correlation coefficient R (the higher the absolute value, the better $\uparrow$) with the corresponding p-value.}
    \label{tab:basket_pear_coef}
\end{table}

We first studied the correlation of each individual evaluation measure to the performance of the policy training separately for the basketball and hockey tasks. In particular, we calculated the Pearson's correlation between the values obtained from each evaluation measure and the corresponding generative model label derived from the performance of the policy as described in Section~\ref{sec:experimental_setup:policy}. We report the Pearson's correlation coefficients (the higher, the better $\uparrow$) together with the corresponding p-values (the lower, the better $\downarrow$), where we consider a correlation to be significant if $p < 0.005$. Furthermore, we studied the same correlation but without differentiating between the tasks. 
Note that our aim was not to train perfect generative models since the variation in the evaluation results among different models enabled us to investigate which of their characteristics are important for the training of GenRL.

\begin{figure}[h]
\centering
\vspace{0.5cm}
\includegraphics[width=0.8\linewidth]{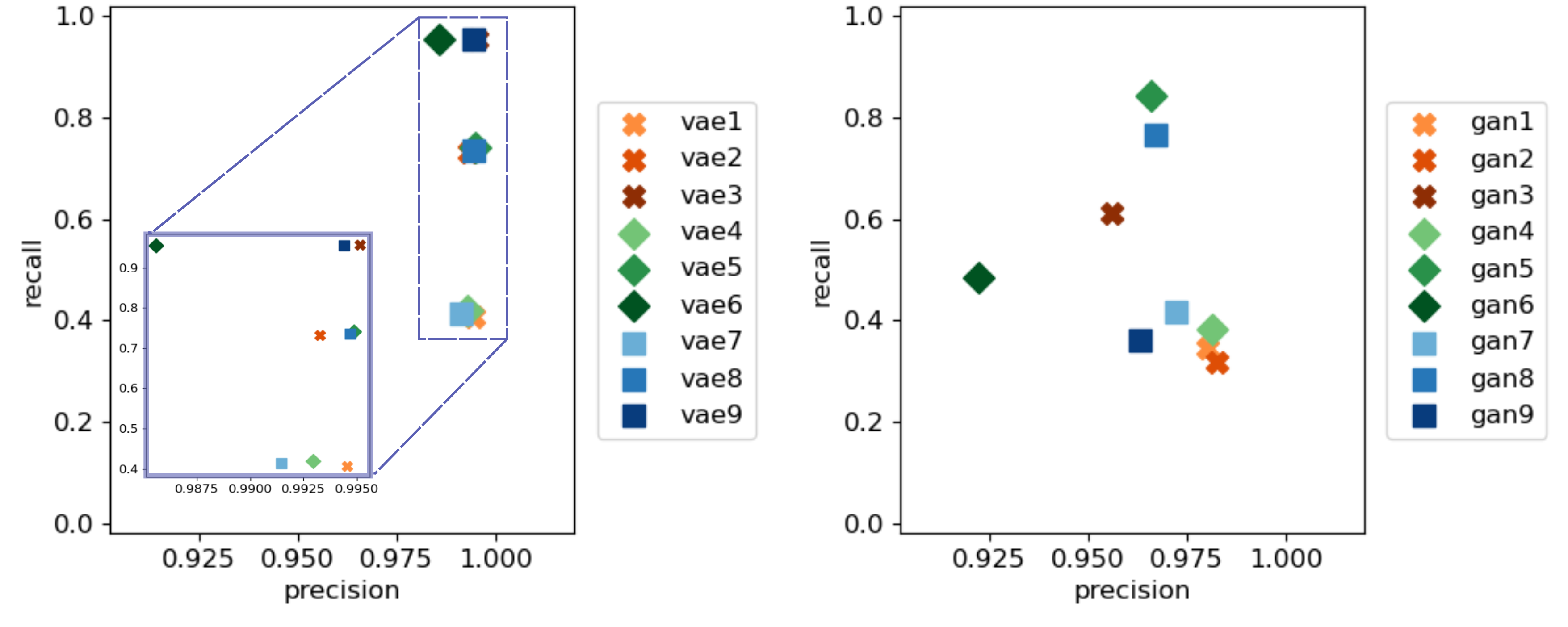} 
\caption{Precision and recall scores for VAE (left) and GAN models (right) trained on the hockey task.}\label{fig:ipr_hockey}
\end{figure}

\begin{figure}[h]
\centering
\includegraphics[width=0.8\linewidth]{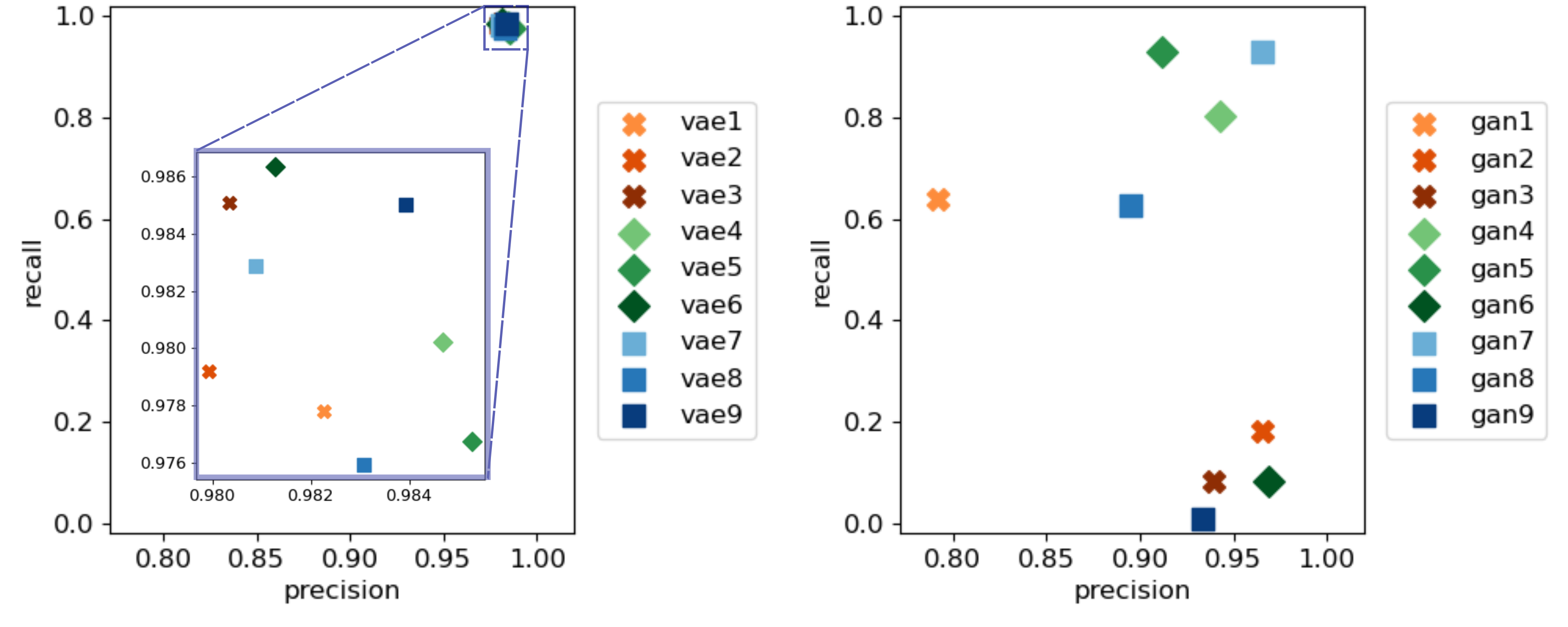} 
\caption{Precision and recall scores for VAE (left) and GAN models (right) trained on the basketball task.}\label{fig:ipr_basket}
\end{figure}

\textbf{Precision and recall} To obtain a successful policy, we expect the generative models to output trajectories that are as similar to the demonstrated motor trajectories as possible. This is measured with precision and recall scores which are shown in Figures~\ref{fig:ipr_hockey} and~\ref{fig:ipr_basket} for the hockey and basketball tasks, respectively. For the \textit{hockey} task (Figure~\ref{fig:ipr_hockey}), we observe that VAEs have higher precision ($x$-axis) and on average higher recall ($y$-axis) than GANs. In Table~\ref{tab:hockey_pear_coef}, we observe a positive correlation of the recall to the policy performance for VAEs, while for GANs neither precision nor recall were found significant. This is likely due to the fact that all GAN models achieve very similar maximum reward (see Table~\ref{tab:models}). However, we observe that GAN5, GAN6 and GAN3, which achieve the highest maximum reward, have the highest recall among the models. Same holds for VAE6, VAE9 and VAE3 models. 

Similarly, for the \textit{basketball} task (Figure~\ref{fig:ipr_basket}), we observe that VAEs have both very high precision and recall, while GANs exhibit lower recall. Especially low recall is obtained for GAN3, GAN6 and GAN9 models which were all trained with $\lambda = 3.5$ which encourages larger disentanglement. These models all achieved worst reward (see Table~\ref{tab:models}). Note that in general the precision of GAN models trained on basketball task is lower than of those trained on the hockey task. In Table~\ref{tab:basket_pear_coef}, we observe that precision is negatively correlated to the policy performance in case of VAEs. We hypothesize that this is a spurious correlation originating from the importance of the high recall. For GANs, we instead observe a positive correlation for recall. When the type of the model was disregarded in the correlation computation (rows ALL in Tables~\ref{tab:hockey_pear_coef} and~\ref{tab:basket_pear_coef}), \textit{we observe recall to be an important characteristic of the generative models since it is positively correlated to the performance of the policy for both tasks.}

\begin{figure}[h]
\vspace{0.5cm}
\centering
\includegraphics[width=0.8\linewidth]{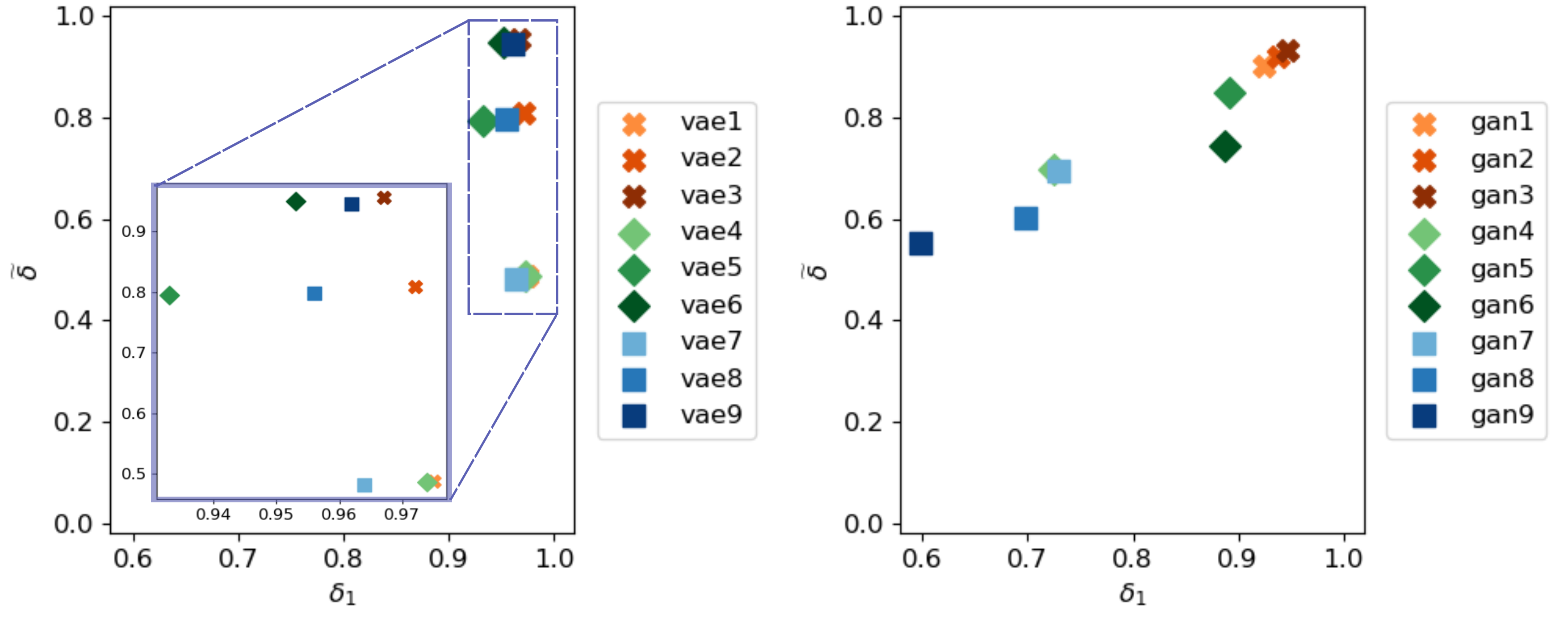} 
\caption{Disentanglement scores for VAE (left) and GAN models (right) trained on the hockey task.}\label{fig:dis_with_ipr_hockey}
\end{figure}

\begin{table}[t]
    \centering
    \vspace{0.2cm}
    \begin{tabular}{r|c|c|c|c|c|c|c|c|c}
    \textit{hockey} & VAE1 & VAE2 & VAE3 & VAE4 & VAE5 & VAE6 & VAE7 & VAE8 & VAE9  \\
     \hline
    \textbf{Precision} & 1.000 & 0.988 & 0.979 & 1.000 & 0.984 & 0.984 & 1.000 & 0.988 & 0.986 \\
    \textbf{Recall} & 0.025 & 0.016 & 0.012 & 0.026 & 0.051 & 0.032 & 0.036 & 0.032 & 0.024 \\
    $\boldsymbol{\delta_1}$ & \cellcolor[gray]{0.65}0.975 & \cellcolor[gray]{0.85}0.972 & 0.967 & \cellcolor[gray]{0.75}0.974 & 0.933 & 0.953 & 0.964 & 0.956 & 0.962 \\

    \hline
    \textbf{Precision} & 1.000 & 0.989 & 0.984 & 1.000 & 0.988 & 0.984 & 1.000 & 0.986 & 0.985 \\
    \textbf{Recall} & 1.000 & 0.344 & 0.04 & 1.000 & 0.332 & 0.037 & 1.000 & 0.346 & 0.055 \\
    $\boldsymbol{\delta_2}$ & 0.000 & 0.645 & \cellcolor[gray]{0.75}0.945 & 0.000 & 0.656 & \cellcolor[gray]{0.65}0.947 & 0.000 & 0.640 & \cellcolor[gray]{0.85}0.930 \\

        \hline
    $\boldsymbol{\widetilde{\delta}}$ & 0.487 & 0.808 & \cellcolor[gray]{0.65}0.956 & 0.487 & 0.794 & \cellcolor[gray]{0.75}0.950 & 0.482 & 0.798 & \cellcolor[gray]{0.85}0.946 \\

    \multicolumn{10}{c}{} \\
    \textit{hockey} & GAN1 & GAN2 & GAN3 & GAN4 & GAN5 & GAN6 & GAN7 & GAN8 & GAN9  \\
    \hline
    \textbf{Precision} & 0.966 & 0.965 & 0.960 & 0.975 & 0.957 & 0.952 & 0.961 & 0.949 & 0.937 \\
    \textbf{Recall} & 0.041 & 0.027 & 0.013 & 0.251 & 0.065 & 0.066 & 0.232 & 0.250 & 0.340 \\
    $\boldsymbol{\delta_1}$ & \cellcolor[gray]{0.85}0.925 & \cellcolor[gray]{0.75}0.938 & \cellcolor[gray]{0.65}0.947 & 0.724 & 0.892 & 0.887 & 0.729 & 0.698 & 0.598 \\

    \hline
    \textbf{Precision} & 0.981 & 0.954 & 0.955 & 0.978 & 0.952 & 0.953 & 0.964 & 0.943 & 0.94 \\
    \textbf{Recall} & 0.103 & 0.051 & 0.034 & 0.304 & 0.145 & 0.348 & 0.303 & 0.434 & 0.434 \\
    $\boldsymbol{\delta_2}$ & \cellcolor[gray]{0.85}0.878 & \cellcolor[gray]{0.75}0.904 & \cellcolor[gray]{0.65}0.921 & 0.674 & 0.807 & 0.604 & 0.661 & 0.509 & 0.506 \\

    \hline
    $\boldsymbol{\widetilde{\delta}}$ & \cellcolor[gray]{0.85}0.901 & \cellcolor[gray]{0.75}0.921 & \cellcolor[gray]{0.65}0.934 & 0.699 & 0.850 & 0.746 & 0.695 & 0.603 & 0.552 \\

    \end{tabular}
    \vspace{0.5cm}
    \caption{Disentanglement scores for VAE (top rows) and GAN (bottom rows) models obtained on the hockey task. We report the precision and recall values calculated as in Equation~\eqref{eq:disentanglement} that yield the highest two values $\delta_1$ and $\delta_2$. We additionally report their average $\widetilde{\delta}$.} 
    \label{tab:hockey_dis_with_pr}
\end{table} 

\begin{figure}[t]
\centering
\includegraphics[width=0.8\linewidth]{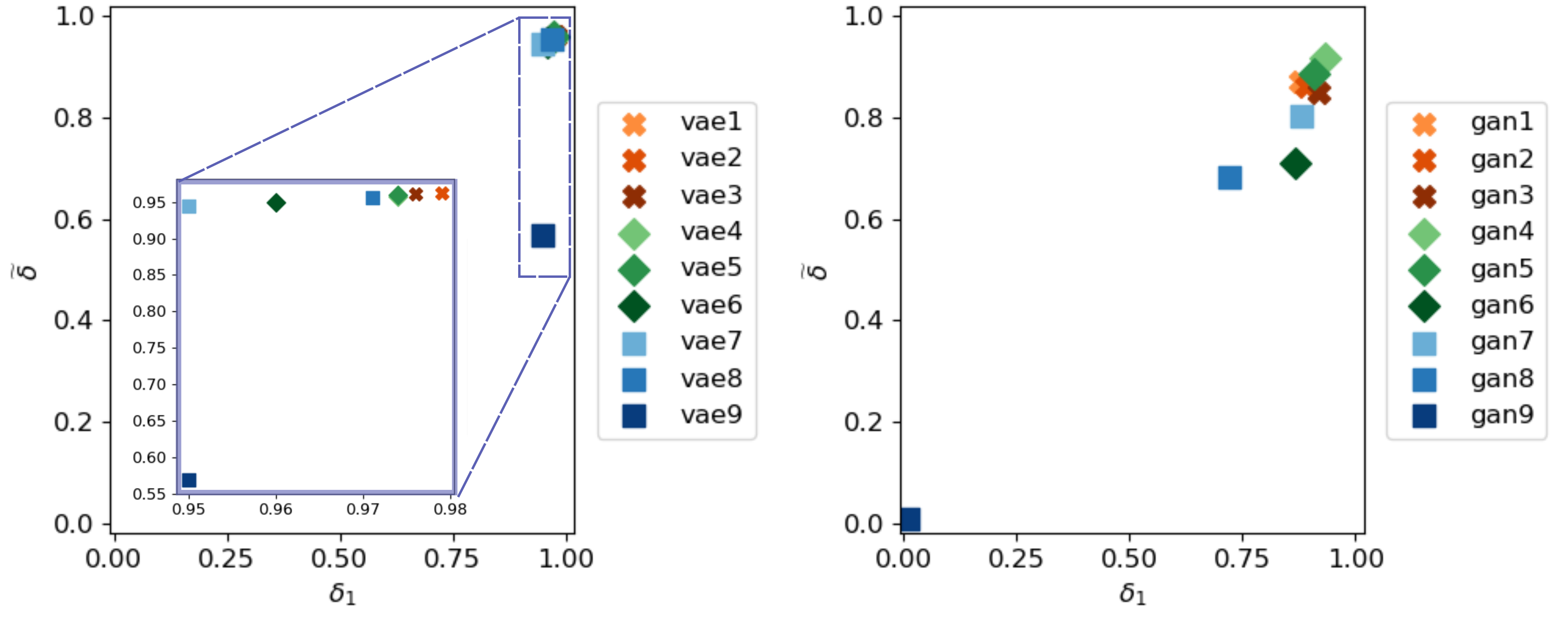} 
\caption{Disentanglement scores for VAE (left) and GAN models (right) trained on the basketball task.}\label{fig:dis_with_ipr_basket}
\end{figure}

\textbf{Disentanglement} Next, we evaluated whether disentangled generative models yield a better policy trained with GenRL. Note again that we considered $\boldsymbol{T}_g^{p(\alpha)}$ as the reference set since we wished to determine how disentangled each model is with respect to its own modelling capabilities. We report the exact disentanglement scores $\delta_1, \widetilde{\delta}$ together with the respective precision and recall values in Tables~\ref{tab:hockey_dis_with_pr} and~\ref{tab:basket_dis_with_pr}, and provide visualization of $\delta_1$ and $\widetilde{\delta}$ in Figures~\ref{fig:dis_with_ipr_hockey} and~\ref{fig:dis_with_ipr_basket} for the hockey and basketball tasks, respectively.

\begin{table}[t]
    \centering
    \vspace{0.2cm}
   
    \begin{tabular}{r|c|c|c|c|c|c|c|c|c}
    \textit{basketball} & VAE1 & VAE2 & VAE3 & VAE4 & VAE5 & VAE6 & VAE7 & VAE8 & VAE9  \\
     \hline
    \textbf{Precision} & 0.983 & 0.988 & 0.985 & 0.989 & 0.984 & 0.974 & 0.984 & 0.991 & 0.986 \\
    \textbf{Recall} & 0.009 & 0.009 & 0.008 & 0.015 & 0.010 & 0.015 & 0.034 & 0.020 & 0.036 \\
    $\boldsymbol{\delta_1}$ &  \cellcolor[gray]{0.85}0.974 & \cellcolor[gray]{0.65} 0.979 & \cellcolor[gray]{0.75} 0.976 & \cellcolor[gray]{0.85}0.974 &  \cellcolor[gray]{0.85}0.974 & 0.960 & 0.950 & 0.971 & 0.950 \\

    \hline
    \textbf{Precision} & 0.986 & 0.986 & 0.986 & 0.979 & 0.984 & 0.983 & 0.985 & 0.987 & 0.984 \\

    \textbf{Recall} & 0.042 & 0.041 & 0.040 & 0.038 & 0.037 & 0.045 & 0.045 & 0.045 & 0.796 \\

    $\boldsymbol{\delta_2}$ & \cellcolor[gray]{0.85}0.944 & \cellcolor[gray]{0.75}0.945 & \cellcolor[gray]{0.65}0.946 & 0.941 & \cellcolor[gray]{0.65}0.946 & 0.938 & 0.940 & 0.942 & 0.189 \\
        \hline
    $\boldsymbol{\widetilde{\delta}}$ & 0.959 & \cellcolor[gray]{0.65}0.962 & \cellcolor[gray]{0.75}0.961 & 0.958 & \cellcolor[gray]{0.85}0.960 & 0.949 & 0.945 & 0.956 & 0.570 \\

    \multicolumn{10}{c}{} \\
    \textit{basketball} & GAN1 & GAN2 & GAN3 & GAN4 & GAN5 & GAN6 & GAN7 & GAN8 & GAN9  \\
    \hline
    \textbf{Precision} & 0.953 & 0.962 & 0.985 & 0.965 & 0.95 & 0.976 & 0.955 & 0.944 & 1.000 \\
    \textbf{Recall} & 0.073 & 0.071 & 0.067 & 0.032 & 0.042 & 0.109 & 0.075 & 0.221 & 0.989 \\

    $\boldsymbol{\delta_1}$ & 0.879 & 0.891 &  \cellcolor[gray]{0.75}0.918 &  \cellcolor[gray]{0.65}0.933 &  \cellcolor[gray]{0.85}0.908 & 0.867 & 0.880 & 0.723 & 0.011 \\

    \hline
    
    \textbf{Precision} & 0.957 & 0.966 & 0.989 & 0.949 & 0.956 & 0.995 & 0.951 & 0.925 & 1.000 \\

    \textbf{Recall} & 0.096 & 0.134 & 0.205 & 0.043 & 0.092 & 0.440 & 0.222 & 0.281 & 0.998 \\

    $\boldsymbol{\delta_2}$ &  \cellcolor[gray]{0.85}0.862 & 0.832 & 0.784 &  \cellcolor[gray]{0.65}0.905 &  \cellcolor[gray]{0.75}0.864 & 0.555 & 0.729 & 0.644 & 0.002 \\
    \hline
    $\boldsymbol{\widetilde{\delta}}$ &  \cellcolor[gray]{0.85}0.871 & 0.861 & 0.851 &  \cellcolor[gray]{0.65}0.919 &  \cellcolor[gray]{0.75}0.886 & 0.711 & 0.804 & 0.683 & 0.006 \\

    \end{tabular}
    \vspace{0.5cm}
    \caption{Disentanglement scores for VAE (top rows) and GAN (bottom rows) models obtained on the basketball task. We report the precision and recall values calculated as in Equation~\eqref{eq:disentanglement} that yield the highest two values $\delta_1$ and $\delta_2$. We additionally report their average $\widetilde{\delta}$.} 
     \label{tab:basket_dis_with_pr}
\end{table} 

For the \textit{hockey} task (Figure~\ref{fig:dis_with_ipr_hockey}), we observe that all generative models attain similar average $\widetilde{\delta}$ of two most disentangled latent dimensions ($y$-axis) and that GAN4, GAN7-9 models (right) attain worse maximum disentanglement $\delta_1$ ($x$-axis) than VAEs (left). These models are trained with larger dimension of the latent space, $N_\alpha = 3 $ and $6$, respectively. We observe that GAN5, GAN6 and GAN3 models, which achieve the best maximum reward, all have both high $\delta_1$ and high $\widetilde{\delta}$. Together with the precision and recall results visualized in Figure~\ref{fig:ipr_hockey}, this suggest that the models necessarily need to have both high recall and be well disentangled. We obtain similar results for VAEs, where we observe a lower average disentanglement $\widetilde{\delta}$ for VAE1, VAE4 and VAE7, which result in worse policy performance as seen in Table~\ref{tab:models}. These models are trained with $\beta = 0.01$ which should learn more disentangled latent representations. However, we observe in Table~\ref{tab:hockey_dis_with_pr} (top) that $\beta = 0.01$ seems to highly disentangle only one latent dimension but ignoring the others. This is on par with observations in Table~\ref{tab:hockey_pear_coef}, where we observe a positive correlation of $\widetilde{\delta}$ to the policy performance for VAEs. Moreover, we observe that VAE3, VAE6 and VAE9 achieve highest $\delta_1$ and $\widetilde{\delta}$, which are the models with highest precision and recall scores and best performing policy. 

For the \textit{basketball} task (Figure~\ref{fig:dis_with_ipr_basket}), we firstly observe that all VAEs achieve high $\delta_1$ and $\widetilde{\delta}$ except for VAE9 (see also Table~\ref{tab:basket_dis_with_pr}) which also achieves one of the lowest maximum reward. Secondly, we observe that GANs obtain lower $\widetilde{\delta}$ than VAEs. We observe that the best performing GAN4, GAN5 and GAN7 achieve high disentanglement values. These models also achieve the highest precision and recall values (see again Figure~\ref{fig:ipr_basket}, right). We observe the worst disentanglement scores for GAN9 which most likely originate from the poor recall of the model. When investigating the correlation of $\delta_1$ and $\widetilde{\delta}$ to the performance of the policy in Table~\ref{tab:basket_pear_coef}, we find that the correlation is not significant when we distinguish the type of the generative models (rows VAE and GAN) due to the dominance of the recall measure. However, when analyzing the correlation without distinguishing the type of the model, we observe a positive correlation for both scores (row ALL).
\textit{Therefore, we conclude that disentanglement is an additionally beneficial characteristic for the generative models that obtain sufficiently high recall. }

\textbf{General evaluation} Lastly, we evaluated the correlation of the evaluation measures to the policy performance for all the models and both tasks. The results, shown in Table~\ref{tab:alltasks_allmodels_pear_coef}, support our earlier conclusions. We observe the highest correlation for recall followed by $\delta_1$. We also observe a weaker positive correlation for precision and $\widetilde{\delta}$. \textit{Therefore, we conclude that the latent action representations should capture all aspects of the robotic task in a disentangled manner, while it is less important to capture them precisely.}

\begin{table}[h]
    \centering
    \vspace{0.2cm}
    
    \begin{tabular}{c|r|c|c|c|c}
    \multicolumn{2}{c|}{\textit{all tasks}}
    & \textbf{Precision} & \textbf{Recall} & $\boldsymbol{\delta_1}$ & $\boldsymbol{\widetilde{\delta}}$ 
    \\
    \hline
    \multirow{2}{*}{ALL} & Pearson's R & \cellcolor[gray]{0.902}$0.479$ &  \cellcolor[gray]{0.75}$0.650$ &  \cellcolor[gray]{0.861}$0.538$ &  \cellcolor[gray]{0.942}$0.458$   \\
    & p-value &  \cellcolor[gray]{0.902}$0.003$ &  \cellcolor[gray]{0.75}$0.000$ &  \cellcolor[gray]{0.861}$0.001$ & \cellcolor[gray]{0.942} $0.005$ \\
    \end{tabular}
    \vspace{0.5cm}
    \caption{Correlation results between the performance of the policy reported in Table~\ref{tab:models} and resulting evaluation measures described in Section \ref{sec:eval_generative_model} for all generative models on both basketball and hockey task. We report Pearson correlation coefficient R (the higher, the better $\uparrow$) with the corresponding p-value.} 
    \label{tab:alltasks_allmodels_pear_coef}
\end{table}

\section{Conclusion}
\label{sec:conclusion}
We presented an RL framework that combined with generative models trains deep visuomotor policies in a data-efficient manner. The generative models are integrated with the RL optimization by introducing a latent variable $\alpha$ that is a low-dimensional representation of a sequence of motor actions. Using the latent action variable $\alpha$, we divided the optimization of the parameters $\Theta$ of the deep visuomotor policy $\pi_\Theta(\tau|s)$
into two parts: optimizing the parameters $\vartheta$ of a generative model $p_\vartheta(\tau|\alpha)$ that generates valid sequences of motor actions, and optimizing the parameters $\theta$ of a sub-policy $\pi_\theta(\alpha | s)$, where $\Theta = [\theta, \vartheta]$. The sub-policy parameters $\theta$ are found using the EM algorithm, while generative model parameters $\vartheta$ are trained unsupervised to optimize the objective corresponding to the chosen generative model. In summary, the complete framework consists of two data-efficient subsequent tasks, training the generative model $p_\vartheta$, and training the sub-policy $\pi_\theta$. 

Moreover, we provided a set of measures for evaluating the quality of the generative models regulated by the RL policy search algorithms such that we can predict the performance of the deep policy training $\pi_\Theta$ prior to the actual training. In particular,
we used the precision and recall measure \cite{kynkaanniemi2019improved} to evaluate the quality of the generated samples, and adjusted it to define a novel measure called disentanglement with precision and recall (DwPR) that additionally evaluates the quality of the latent space of the generative model $p_\vartheta$.
We experimentally demonstrated the predictive power of these measures on two tasks in simulation, shooting hockey puck and throwing a basketball, using a set of different VAE and GAN generative models. 
Regardless of the model type, we observe that the most beneficial latent action representations are those that capture all aspects of the robotic task in a disentangled manner.

\section*{Acknowledgments}
This work was supported by Knut and Alice Wallenberg Foundation, the EU through the project EnTimeMent, the Swedish Foundation for Strategic Research through the COIN project, and also by the  Academy of Finland through the DEEPEN project.

\bibliographystyle{IEEEtran}
\bibliography{ref}

\begin{thebibliography}{10}
\providecommand{\url}[1]{#1}
\csname url@samestyle\endcsname
\providecommand{\newblock}{\relax}
\providecommand{\bibinfo}[2]{#2}
\providecommand{\BIBentrySTDinterwordspacing}{\spaceskip=0pt\relax}
\providecommand{\BIBentryALTinterwordstretchfactor}{4}
\providecommand{\BIBentryALTinterwordspacing}{\spaceskip=\fontdimen2\font plus
\BIBentryALTinterwordstretchfactor\fontdimen3\font minus
  \fontdimen4\font\relax}
\providecommand{\BIBforeignlanguage}[2]{{%
\expandafter\ifx\csname l@#1\endcsname\relax
\typeout{** WARNING: IEEEtran.bst: No hyphenation pattern has been}%
\typeout{** loaded for the language `#1'. Using the pattern for}%
\typeout{** the default language instead.}%
\else
\language=\csname l@#1\endcsname
\fi
#2}}
\providecommand{\BIBdecl}{\relax}
\BIBdecl

\bibitem{singh2020parrot}
A.~Singh, H.~Liu, G.~Zhou, A.~Yu, N.~Rhinehart, and S.~Levine, ``Parrot:
  Data-driven behavioral priors for reinforcement learning,'' in
  \emph{International Conference on Learning Representations}, 2020.

\bibitem{ghadirzadeh2017deep}
A.~Ghadirzadeh, A.~Maki, D.~Kragic, and M.~Bj{\"o}rkman, ``Deep predictive
  policy training using reinforcement learning,'' in \emph{2017 IEEE/RSJ
  International Conference on Intelligent Robots and Systems (IROS)}.\hskip 1em
  plus 0.5em minus 0.4em\relax IEEE, 2017, pp. 2351--2358.

\bibitem{arndt2019meta}
K.~Arndt, M.~Hazara, A.~Ghadirzadeh, and V.~Kyrki, ``Meta reinforcement
  learning for sim-to-real domain adaptation,'' in \emph{2020 IEEE
  International Conference on Robotics and Automation (ICRA)}, 2020.

\bibitem{levine2016end}
S.~Levine, C.~Finn, T.~Darrell, and P.~Abbeel, ``End-to-end training of deep
  visuomotor policies,'' \emph{The Journal of Machine Learning Research},
  vol.~17, no.~1, pp. 1334--1373, 2016.

\bibitem{levine2018learning}
S.~Levine, P.~Pastor, A.~Krizhevsky, J.~Ibarz, and D.~Quillen, ``Learning
  hand-eye coordination for robotic grasping with deep learning and large-scale
  data collection,'' \emph{The International Journal of Robotics Research},
  vol.~37, no. 4-5, pp. 421--436, 2018.

\bibitem{kynkaanniemi2019improved}
T.~Kynk{\"a}{\"a}nniemi, T.~Karras, S.~Laine, J.~Lehtinen, and T.~Aila,
  ``Improved precision and recall metric for assessing generative models,'' in
  \emph{Advances in Neural Information Processing Systems}, 2019, pp.
  3927--3936.

\bibitem{higgins2017beta}
I.~Higgins, L.~Matthey, A.~Pal, C.~Burgess, X.~Glorot, M.~Botvinick,
  S.~Mohamed, and A.~Lerchner, ``beta-vae: Learning basic visual concepts with
  a constrained variational framework,'' in \emph{International Conference on
  Learning Representations}, 2017.

\bibitem{chen2016infogan}
X.~Chen, Y.~Duan, R.~Houthooft, J.~Schulman, I.~Sutskever, and P.~Abbeel,
  ``Infogan: Interpretable representation learning by information maximizing
  generative adversarial nets,'' in \emph{Advances in neural information
  processing systems}, 2016, pp. 2172--2180.

\bibitem{schulman2017proximal}
J.~Schulman, F.~Wolski, P.~Dhariwal, A.~Radford, and O.~Klimov, ``Proximal
  policy optimization algorithms,'' \emph{arXiv preprint arXiv:1707.06347},
  2017.

\bibitem{haarnoja2018soft}
T.~Haarnoja, A.~Zhou, P.~Abbeel, and S.~Levine, ``Soft actor-critic: Off-policy
  maximum entropy deep reinforcement learning with a stochastic actor,'' in
  \emph{International conference on machine learning}.\hskip 1em plus 0.5em
  minus 0.4em\relax PMLR, 2018, pp. 1861--1870.

\bibitem{chen2019adversarial}
X.~Chen, A.~Ghadirzadeh, M.~Bj{\"o}rkman, and P.~Jensfelt, ``Adversarial
  feature training for generalizable robotic visuomotor control,'' in
  \emph{2020 IEEE International Conference on Robotics and Automation (ICRA)},
  2020.

\bibitem{hamalainen2019affordance}
A.~H{\"a}m{\"a}l{\"a}inen, K.~Arndt, A.~Ghadirzadeh, and V.~Kyrki, ``Affordance
  learning for end-to-end visuomotor robot control,'' in \emph{2019 IEEE/RSJ
  international conference on intelligent robots and systems (IROS)}, 2019.

\bibitem{butepage2019imitating}
J.~B{\"u}tepage, A.~Ghadirzadeh, {\"O}.~{\"O}. Karadag, M.~Bj{\"o}rkman, and
  D.~Kragic, ``Imitating by generating: deep generative models for imitation of
  interactive tasks,'' \emph{Frontiers in Robotics and AI}, 2020.

\bibitem{finn2016deep}
C.~Finn, X.~Y. Tan, Y.~Duan, T.~Darrell, S.~Levine, and P.~Abbeel, ``Deep
  spatial autoencoders for visuomotor learning,'' in \emph{2016 IEEE
  International Conference on Robotics and Automation (ICRA)}.\hskip 1em plus
  0.5em minus 0.4em\relax IEEE, 2016, pp. 512--519.

\bibitem{kalashnikov2018qt}
D.~Kalashnikov, A.~Irpan, P.~Pastor, J.~Ibarz, A.~Herzog, E.~Jang, D.~Quillen,
  E.~Holly, M.~Kalakrishnan, V.~Vanhoucke, and S.~Levine, ``Qt-opt: Scalable
  deep reinforcement learning for vision-based robotic manipulation,'' in
  \emph{2nd Conference on Robot Learning (CoRL)}, 2018.

\bibitem{quillen2018deep}
D.~Quillen, E.~Jang, O.~Nachum, C.~Finn, J.~Ibarz, and S.~Levine, ``Deep
  reinforcement learning for vision-based robotic grasping: A simulated
  comparative evaluation of off-policy methods,'' in \emph{2018 IEEE
  International Conference on Robotics and Automation (ICRA)}.\hskip 1em plus
  0.5em minus 0.4em\relax IEEE, 2018, pp. 6284--6291.

\bibitem{singh2017gplac}
A.~Singh, L.~Yang, and S.~Levine, ``Gplac: Generalizing vision-based robotic
  skills using weakly labeled images,'' in \emph{Proceedings of the IEEE
  International Conference on Computer Vision}, 2017, pp. 5851--5860.

\bibitem{devin2018deep}
C.~Devin, P.~Abbeel, T.~Darrell, and S.~Levine, ``Deep object-centric
  representations for generalizable robot learning,'' in \emph{2018 IEEE
  International Conference on Robotics and Automation (ICRA)}.\hskip 1em plus
  0.5em minus 0.4em\relax IEEE, 2018, pp. 7111--7118.

\bibitem{pinto2017asymmetric}
L.~Pinto, M.~Andrychowicz, P.~Welinder, W.~Zaremba, and P.~Abbeel, ``Asymmetric
  actor critic for image-based robot learning,'' \emph{arXiv preprint
  arXiv:1710.06542}, 2017.

\bibitem{finn2017deep}
C.~Finn and S.~Levine, ``Deep visual foresight for planning robot motion,'' in
  \emph{2017 IEEE International Conference on Robotics and Automation
  (ICRA)}.\hskip 1em plus 0.5em minus 0.4em\relax IEEE, 2017, pp. 2786--2793.

\bibitem{gu2017deep}
S.~Gu, E.~Holly, T.~Lillicrap, and S.~Levine, ``Deep reinforcement learning for
  robotic manipulation with asynchronous off-policy updates,'' in \emph{2017
  IEEE international conference on robotics and automation (ICRA)}.\hskip 1em
  plus 0.5em minus 0.4em\relax IEEE, 2017, pp. 3389--3396.

\bibitem{dasari2019robonet}
S.~Dasari, F.~Ebert, S.~Tian, S.~Nair, B.~Bucher, K.~Schmeckpeper, S.~Singh,
  S.~Levine, and C.~Finn, ``Robonet: Large-scale multi-robot learning,''
  \emph{arXiv preprint arXiv:1910.11215}, 2019.

\bibitem{abdolmaleki2020distributional}
A.~Abdolmaleki, S.~H. Huang, L.~Hasenclever, M.~Neunert, H.~F. Song,
  M.~Zambelli, M.~F. Martins, N.~Heess, R.~Hadsell, and M.~Riedmiller, ``A
  distributional view on multi-objective policy optimization,'' \emph{arXiv
  preprint arXiv:2005.07513}, 2020.

\bibitem{peng2018sim}
X.~B. Peng, M.~Andrychowicz, W.~Zaremba, and P.~Abbeel, ``Sim-to-real transfer
  of robotic control with dynamics randomization,'' in \emph{2018 IEEE
  international conference on robotics and automation (ICRA)}.\hskip 1em plus
  0.5em minus 0.4em\relax IEEE, 2018, pp. 1--8.

\bibitem{tobin2017domain}
J.~Tobin, R.~Fong, A.~Ray, J.~Schneider, W.~Zaremba, and P.~Abbeel, ``Domain
  randomization for transferring deep neural networks from simulation to the
  real world,'' in \emph{2017 IEEE/RSJ international conference on intelligent
  robots and systems (IROS)}.\hskip 1em plus 0.5em minus 0.4em\relax IEEE,
  2017, pp. 23--30.

\bibitem{yu2018one}
T.~Yu, C.~Finn, A.~Xie, S.~Dasari, T.~Zhang, P.~Abbeel, and S.~Levine,
  ``One-shot imitation from observing humans via domain-adaptive
  meta-learning,'' \emph{arXiv preprint arXiv:1802.01557}, 2018.

\bibitem{chen2018deep}
X.~Chen, A.~Ghadirzadeh, J.~Folkesson, M.~Bj{\"o}rkman, and P.~Jensfelt, ``Deep
  reinforcement learning to acquire navigation skills for wheel-legged robots
  in complex environments,'' in \emph{2018 IEEE/RSJ International Conference on
  Intelligent Robots and Systems (IROS)}.\hskip 1em plus 0.5em minus
  0.4em\relax IEEE, 2018, pp. 3110--3116.

\bibitem{tzeng2017adversarial}
E.~Tzeng, J.~Hoffman, K.~Saenko, and T.~Darrell, ``Adversarial discriminative
  domain adaptation,'' in \emph{Proceedings of the IEEE Conference on Computer
  Vision and Pattern Recognition}, 2017, pp. 7167--7176.

\bibitem{tzeng2020adapting}
E.~Tzeng, C.~Devin, J.~Hoffman, C.~Finn, P.~Abbeel, S.~Levine, K.~Saenko, and
  T.~Darrell, ``Adapting deep visuomotor representations with weak pairwise
  constraints,'' in \emph{Algorithmic Foundations of Robotics XII}.\hskip 1em
  plus 0.5em minus 0.4em\relax Springer, 2020, pp. 688--703.

\bibitem{schulman2015trust}
J.~Schulman, S.~Levine, P.~Abbeel, M.~Jordan, and P.~Moritz, ``Trust region
  policy optimization,'' in \emph{International conference on machine
  learning}, 2015, pp. 1889--1897.

\bibitem{neumann2011variational}
G.~Neumann, ``Variational inference for policy search in changing situations,''
  in \emph{Proceedings of the 28th International Conference on Machine
  Learning, ICML 2011}, 2011, pp. 817--824.

\bibitem{deisenroth2013survey}
M.~P. Deisenroth, G.~Neumann, J.~Peters \emph{et~al.}, ``A survey on policy
  search for robotics,'' \emph{Foundations and Trends{\textregistered} in
  Robotics}, vol.~2, no. 1--2, pp. 1--142, 2013.

\bibitem{levine2013variational}
S.~Levine and V.~Koltun, ``Variational policy search via trajectory
  optimization,'' in \emph{Advances in neural information processing systems},
  2013, pp. 207--215.

\bibitem{ghadirzadeh2018sensorimotor}
A.~Ghadirzadeh, ``Sensorimotor robot policy training using reinforcement
  learning,'' Ph.D. dissertation, KTH Royal Institute of Technology, 2018.

\bibitem{bahl2020neural}
S.~Bahl, M.~Mukadam, A.~Gupta, and D.~Pathak, ``Neural dynamic policies for
  end-to-end sensorimotor learning,'' \emph{Advances in Neural Information
  Processing Systems}, vol.~33, pp. 5058--5069, 2020.

\bibitem{peters2006policy}
J.~Peters and S.~Schaal, ``Policy gradient methods for robotics,'' in
  \emph{2006 IEEE/RSJ International Conference on Intelligent Robots and
  Systems}.\hskip 1em plus 0.5em minus 0.4em\relax IEEE, 2006, pp. 2219--2225.

\bibitem{peters2008reinforcement}
------, ``Reinforcement learning of motor skills with policy gradients,''
  \emph{Neural networks}, vol.~21, no.~4, pp. 682--697, 2008.

\bibitem{ijspeert2003learning}
A.~J. Ijspeert, J.~Nakanishi, and S.~Schaal, ``Learning attractor landscapes
  for learning motor primitives,'' in \emph{Advances in neural information
  processing systems}, 2003, pp. 1547--1554.

\bibitem{ijspeert2013dynamical}
A.~J. Ijspeert, J.~Nakanishi, H.~Hoffmann, P.~Pastor, and S.~Schaal,
  ``Dynamical movement primitives: learning attractor models for motor
  behaviors,'' \emph{Neural computation}, vol.~25, no.~2, pp. 328--373, 2013.

\bibitem{hazara2019transferring}
M.~Hazara and V.~Kyrki, ``Transferring generalizable motor primitives from
  simulation to real world,'' \emph{IEEE Robotics and Automation Letters},
  vol.~4, no.~2, pp. 2172--2179, 2019.

\bibitem{haarnoja2018composable}
T.~Haarnoja, V.~Pong, A.~Zhou, M.~Dalal, P.~Abbeel, and S.~Levine, ``Composable
  deep reinforcement learning for robotic manipulation,'' in \emph{2018 IEEE
  International Conference on Robotics and Automation (ICRA)}.\hskip 1em plus
  0.5em minus 0.4em\relax IEEE, 2018, pp. 6244--6251.

\bibitem{zhou2020plas}
W.~Zhou, S.~Bajracharya, and D.~Held, ``Plas: Latent action space for offline
  reinforcement learning,'' \emph{arXiv preprint arXiv:2011.07213}, 2020.

\bibitem{chen2022latent}
X.~Chen, A.~Ghadirzadeh, T.~Yu, Y.~Gao, J.~Wang, W.~Li, B.~Liang, C.~Finn, and
  C.~Zhang, ``Latent-variable advantage-weighted policy optimization for
  offline rl,'' \emph{arXiv preprint arXiv:2203.08949}, 2022.

\bibitem{lippi2020latent}
M.~Lippi, P.~Poklukar, M.~C. Welle, A.~Varava, H.~Yin, A.~Marino, and
  D.~Kragic, ``Latent space roadmap for visual action planning of deformable
  and rigid object manipulation,'' \emph{arXiv preprint arXiv:2003.08974},
  2020.

\bibitem{gothoskar2020learning}
N.~Gothoskar, M.~L{\'a}zaro-Gredilla, A.~Agarwal, Y.~Bekiroglu, and D.~George,
  ``Learning a generative model for robot control using visual feedback,''
  \emph{arXiv preprint arXiv:2003.04474}, 2020.

\bibitem{igl2018deep}
M.~Igl, L.~Zintgraf, T.~A. Le, F.~Wood, and S.~Whiteson, ``Deep variational
  reinforcement learning for pomdps,'' \emph{arXiv preprint arXiv:1806.02426},
  2018.

\bibitem{buesing2018learning}
L.~Buesing, T.~Weber, S.~Racaniere, S.~Eslami, D.~Rezende, D.~P. Reichert,
  F.~Viola, F.~Besse, K.~Gregor, D.~Hassabis \emph{et~al.}, ``Learning and
  querying fast generative models for reinforcement learning,'' \emph{arXiv
  preprint arXiv:1802.03006}, 2018.

\bibitem{mishra2017prediction}
N.~Mishra, P.~Abbeel, and I.~Mordatch, ``Prediction and control with temporal
  segment models,'' in \emph{Proceedings of the 34th International Conference
  on Machine Learning-Volume 70}.\hskip 1em plus 0.5em minus 0.4em\relax JMLR.
  org, 2017, pp. 2459--2468.

\bibitem{ke2018modeling}
N.~R. Ke, A.~Singh, A.~Touati, A.~Goyal, Y.~Bengio, D.~Parikh, and D.~Batra,
  ``Modeling the long term future in model-based reinforcement learning,'' in
  \emph{International Conference on Learning Representations}, 2018.

\bibitem{hafner2018learning}
D.~Hafner, T.~Lillicrap, I.~Fischer, R.~Villegas, D.~Ha, H.~Lee, and
  J.~Davidson, ``Learning latent dynamics for planning from pixels,''
  \emph{arXiv preprint arXiv:1811.04551}, 2018.

\bibitem{rhinehart2018deep}
N.~Rhinehart, R.~McAllister, and S.~Levine, ``Deep imitative models for
  flexible inference, planning, and control,'' \emph{arXiv preprint
  arXiv:1810.06544}, 2018.

\bibitem{krupnik2019multi}
O.~Krupnik, I.~Mordatch, and A.~Tamar, ``Multi agent reinforcement learning
  with multi-step generative models,'' \emph{arXiv preprint arXiv:1901.10251},
  2019.

\bibitem{dinh2016density}
L.~Dinh, J.~Sohl-Dickstein, and S.~Bengio, ``Density estimation using real
  nvp,'' \emph{arXiv preprint arXiv:1605.08803}, 2016.

\bibitem{brock2018large}
A.~Brock, J.~Donahue, and K.~Simonyan, ``Large scale gan training for high
  fidelity natural image synthesis,'' \emph{arXiv preprint arXiv:1809.11096},
  2018.

\bibitem{wang2018high}
T.-C. Wang, M.-Y. Liu, J.-Y. Zhu, A.~Tao, J.~Kautz, and B.~Catanzaro,
  ``High-resolution image synthesis and semantic manipulation with conditional
  gans,'' in \emph{Proceedings of the IEEE conference on computer vision and
  pattern recognition}, 2018, pp. 8798--8807.

\bibitem{vae_anom}
F.~{Wiewel} and B.~{Yang}, ``Continual learning for anomaly detection with
  variational autoencoder,'' in \emph{ICASSP 2019 - 2019 IEEE International
  Conference on Acoustics, Speech and Signal Processing (ICASSP)}, 2019, pp.
  3837--3841.

\bibitem{vae_text8672806}
W.~{Xu} and Y.~{Tan}, ``Semisupervised text classification by variational
  autoencoder,'' \emph{IEEE Transactions on Neural Networks and Learning
  Systems}, vol.~31, no.~1, pp. 295--308, 2020.

\bibitem{challenging_common}
F.~Locatello, S.~Bauer, M.~Lučić, G.~Rätsch, S.~Gelly, B.~Schölkopf, and
  O.~F. Bachem, ``Challenging common assumptions in the unsupervised learning
  of disentangled representations,'' in \emph{International Conference on
  Machine Learning}, 2019, best Paper Award.

\bibitem{poklukar2022delaunay}
\BIBentryALTinterwordspacing
P.~Poklukar, V.~Polianskii, A.~Varava, F.~T. Pokorny, and D.~K. Jensfelt,
  ``Delaunay component analysis for evaluation of data representations,'' in
  \emph{International Conference on Learning Representations}, 2022. [Online].
  Available: \url{https://openreview.net/forum?id=HTVch9AMPa}
\BIBentrySTDinterwordspacing

\bibitem{pmlr-v139-poklukar21a}
\BIBentryALTinterwordspacing
P.~Poklukar, A.~Varava, and D.~Kragic, ``Geomca: Geometric evaluation of data
  representations,'' in \emph{Proceedings of the 38th International Conference
  on Machine Learning}, ser. Proceedings of Machine Learning Research, M.~Meila
  and T.~Zhang, Eds., vol. 139.\hskip 1em plus 0.5em minus 0.4em\relax PMLR,
  18--24 Jul 2021, pp. 8588--8598. [Online]. Available:
  \url{https://proceedings.mlr.press/v139/poklukar21a.html}
\BIBentrySTDinterwordspacing

\bibitem{IS_NIPS2016_6125}
T.~Salimans, I.~Goodfellow, W.~Zaremba, V.~Cheung, A.~Radford, X.~Chen, and
  X.~Chen, ``Improved techniques for training gans,'' in \emph{Advances in
  Neural Information Processing Systems 29}, D.~D. Lee, M.~Sugiyama, U.~V.
  Luxburg, I.~Guyon, and R.~Garnett, Eds.\hskip 1em plus 0.5em minus
  0.4em\relax Curran Associates, Inc., 2016, pp. 2234--2242.

\bibitem{FID_NIPS2017_7240}
M.~Heusel, H.~Ramsauer, T.~Unterthiner, B.~Nessler, and S.~Hochreiter, ``Gans
  trained by a two time-scale update rule converge to a local nash
  equilibrium,'' in \emph{Advances in Neural Information Processing Systems
  30}.\hskip 1em plus 0.5em minus 0.4em\relax Curran Associates, Inc., 2017,
  pp. 6626--6637.

\bibitem{binkowski2018demystifying}
M.~Bińkowski, D.~J. Sutherland, M.~Arbel, and A.~Gretton, ``Demystifying {MMD}
  {GAN}s,'' in \emph{International Conference on Learning Representations},
  2018.

\bibitem{sajjadi2018assessing}
M.~S. Sajjadi, O.~Bachem, M.~Lucic, O.~Bousquet, and S.~Gelly, ``Assessing
  generative models via precision and recall,'' in \emph{Advances in Neural
  Information Processing Systems}, 2018, pp. 5228--5237.

\bibitem{revisiting_pr}
L.~Simon, R.~Webster, and J.~Rabin, ``Revisiting precision recall definition
  for generative modeling,'' in \emph{Proceedings of the 36th International
  Conference on Machine Learning}, ser. Proceedings of Machine Learning
  Research, K.~Chaudhuri and R.~Salakhutdinov, Eds., vol.~97.\hskip 1em plus
  0.5em minus 0.4em\relax Long Beach, California, USA: PMLR, 09--15 Jun 2019,
  pp. 5799--5808.

\bibitem{higgins2018towards}
I.~Higgins, D.~Amos, D.~Pfau, S.~Racaniere, L.~Matthey, D.~Rezende, and
  A.~Lerchner, ``Towards a definition of disentangled representations,''
  \emph{arXiv preprint arXiv:1812.02230}, 2018.

\bibitem{repr_learning_survey}
Y.~{Bengio}, A.~{Courville}, and P.~{Vincent}, ``Representation learning: A
  review and new perspectives,'' \emph{IEEE Transactions on Pattern Analysis
  and Machine Intelligence}, vol.~35, no.~8, pp. 1798--1828, 2013.

\bibitem{tschannen2018recent}
M.~Tschannen, O.~Bachem, and M.~Lucic, ``Recent advances in autoencoder-based
  representation learning,'' \emph{arXiv preprint arXiv:1812.05069}, 2018.

\bibitem{kim2018disentangling}
H.~Kim and A.~Mnih, ``Disentangling by factorising,'' \emph{arXiv preprint
  arXiv:1802.05983}, 2018.

\bibitem{eastwood2018framework}
C.~Eastwood and C.~K. Williams, ``A framework for the quantitative evaluation
  of disentangled representations,'' in \emph{International Conference on
  Learning Representations}, 2018.

\bibitem{chen2018isolating}
T.~Q. Chen, X.~Li, R.~B. Grosse, and D.~K. Duvenaud, ``Isolating sources of
  disentanglement in variational autoencoders,'' in \emph{Advances in Neural
  Information Processing Systems}, 2018, pp. 2610--2620.

\bibitem{kumar2017variational}
A.~Kumar, P.~Sattigeri, and A.~Balakrishnan, ``Variational inference of
  disentangled latent concepts from unlabeled observations,'' \emph{arXiv
  preprint arXiv:1711.00848}, 2017.

\bibitem{jeon2019ibgan}
I.~Jeon, W.~Lee, and G.~Kim, ``{IB}-{GAN}: Disentangled representation learning
  with information bottleneck {GAN},'' 2019.

\bibitem{lee2020high}
W.~Lee, D.~Kim, S.~Hong, and H.~Lee, ``High-fidelity synthesis with
  disentangled representation,'' \emph{arXiv preprint arXiv:2001.04296}, 2020.

\bibitem{liu2019oogan}
B.~Liu, Y.~Zhu, Z.~Fu, G.~de~Melo, and A.~Elgammal, ``Oogan: Disentangling gan
  with one-hot sampling and orthogonal regularization,'' 2019.

\bibitem{kingma2014auto}
D.~P. Kingma and M.~Welling, ``Auto-encoding variational bayes,''
  \emph{International Conference on Learning Representations}, 2014.

\bibitem{rezende2014stochasticvae2}
D.~J. Rezende, S.~Mohamed, and D.~Wierstra, ``Stochastic backpropagation and
  approximate inference in deep generative models,'' in \emph{Int. Conf. Mach.
  Learn.}, 2014, pp. 1278--1286.

\bibitem{goodfellow2014generative}
I.~Goodfellow, J.~Pouget-Abadie, M.~Mirza, B.~Xu, D.~Warde-Farley, S.~Ozair,
  A.~Courville, and Y.~Bengio, ``Generative adversarial nets,'' in
  \emph{Advances in neural information processing systems}, 2014, pp.
  2672--2680.

\end{thebibliography}

\newpage
\appendix
\begin{appendices}

\section{End-to-end Training of Perception and Control}
\label{app:perception}
The EM policy training algorithm presented in Section~\ref{sec:em_policy_training} updates the deep policy using the supervised learning objective function introduced in~\eqref{eq:M_loss} (the M-step objective). 
Similar to GPS \cite{levine2016end}, the EM policy training formulation enables simultaneous training of the perception and control parts of the deep policy in an end-to-end fashion. 
In this section, we describe two techniques that can improve the efficiency of the end-to-end training.

\textbf{Input remapping trick}
The input remapping trick \cite{levine2016end} can be applied to condition the variational policy $q$ on a low-dimensional compact state representation, $z$, instead of the high-dimensional states $s$ given by the sensory observations, e.g., camera images. 
The policy training phase can be done in a controlled environment such that extra measures other than the sensory observation of the system can be provided. These extra measures can be for example the position of a target object on a tabletop. 
Therefore, the image observations $s$ can be paired with a compact task-specific state representation $z$ such that $z$ is used in the E-step for updating the variational policy $q_\phi(\alpha|z)$, and $s$ in the M-step for updating the policy $\pi_\theta(\alpha|s)$. 

\textbf{Domain adaptation for perception training}
Domain adaptation techniques, e.g., adversarial methods \cite{chen2019adversarial}, can improve the end-to-end training of visuomotor policies with limited robot data samples. 
The unlabeled task-specific images, captured without involving the robot, can be exploited in the M-step to improve the generality of the visuomotor policy to manipulate novel task objects in cluttered backgrounds. 

The M-step is updated to include an extra loss function to adapt data from the two different domains: (i) unlabeled images and (ii) robot visuomotor data. 
The images must contain only one task object in a cluttered background, possibly different than the task object used by the robot during the policy training. 
Given images from the two domains, the basic idea is to extract visual features such that it is not possible to detect the source of the features. 
More details of the method can be found in our recent work in \cite{chen2019adversarial}.

\begin{table}[!htb]
    \begin{minipage}{.4\linewidth}
    \centering
        \begin{tabular}{l}
         \hline
         \hline
         Lin($N_\alpha$, $128$) + BN + ReLU \\ 
         \hline
         Lin($128$, $256$) + BN + ReLU \\
         \hline
         Lin($256$, $512$) + BN + ReLU \\
         \hline
         Lin($512$, InputDim)
    \end{tabular}
    \vspace{0.5cm}
    \caption{Architecture of the generator neural network.} 
    \label{tab:gen_arc}
    \end{minipage}%
    \hfill
    \begin{minipage}{.55\linewidth}
      \centering
        \begin{tabular}{l|l}
         \hline
         \hline
         \multirow{2}{*}{Shared layers} & Lin(InputDim, $256$) + ReLU \\ 
         \cline{2-2}
         & Lin($256$, $128$) + ReLU \\
         \hline
         discriminator & Lin($128$, $1$) + Sigmoid \\
         \hline \multirow{2}{*}{Qnet} & Lin($128$, $64$) \\ \cline{2-2}
          & Lin($64$, $N_\alpha$) 
         \end{tabular}
         \vspace{0.5cm}
         \caption{Architecture of the discriminator and Qnet neural networks.}
         \label{tab:dis_arc}
    \end{minipage} 
    \vspace{-0.8cm}
\end{table}

\section{Training details of generative models} \label{app:gen_models_details}
In this section, we report architectural and training details for generative models used in our experiments. The architectures of the decoders is shown in Table~\ref{tab:gen_arc}, where we used Lin and BN to refer to the linear and batch normalization layers, respectively. The InputDim denotes the dimension of the trajectories and is equal to $69\cdot 7$ and $78 \cdot 7$ for the hockey and basketball tasks, respectively. %
Architectural choices related to each type of generative model are discussed below. All the models were trained for $5000$ epochs with learning rate fixed to $2e-4$ using batch size $256$. For our evaluation, we chose the best performing models based on minimum squared error computed between $100$ randomly generated trajectories and trajectories in the test datasets not used during training.

\noindent \textbf{Variational Autoencoder} The encoder neural network is symmetric to the decoder with two output linear layers of size $N_\alpha$ representing the mean and the log standard deviation of the approximate posterior distribution. 

\noindent \textbf{InfoGAN} The architecture of the generator, discriminator and the neural network parametrizing $Q_\phi(\alpha|\tau)$ are summarised in Tables~\ref{tab:gen_arc} and~\ref{tab:dis_arc}. Learning rates of the optimizers for the generator and discriminator networks were fixed to $2e-4$. 

\section{Examples of generated trajectories} \label{app:gen_models_traj}
\begin{figure}[t]
\vspace{0.5cm}
\centering
\includegraphics[width=0.95\linewidth]{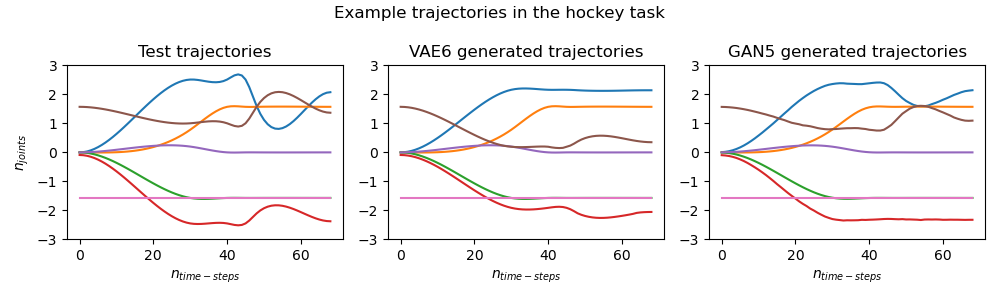}
\caption{Example of a test trajectory (left), VAE6 generated trajectory (middle) and GAN5 generated trajectory (right) for the hockey task.}
\label{fig:examples_hockey}
\vspace{-1.5cm}
\end{figure}

\begin{figure}[t]
\centering
\includegraphics[width=0.95\linewidth]{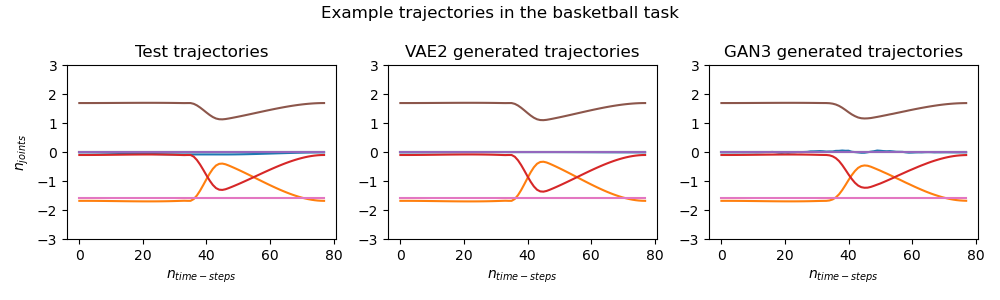}
\caption{Example of a test trajectory (left), VAE2 generated trajectory (middle) and GAN3 generated trajectory (right) for the basketball task.}
\label{fig:examples_basket}

\end{figure}

In this section, we visualize examples of testing trajectories as well as generated ones for both hockey (Figure~\ref{fig:examples_hockey}) and basketball tasks (Figure~\ref{fig:examples_basket}).

\end{appendices}

\end{document}